\documentclass[5p]{elsarticle}
\biboptions{authoryear}

\journal{Medical Image Analysis}

\usepackage{gensymb}
\usepackage{graphicx}
\usepackage[utf8]{inputenc}
\usepackage[normalem]{ulem}

\usepackage{booktabs, multicol, multirow}
\usepackage[table]{xcolor}
\usepackage[T1]{fontenc}

\usepackage{hyperref}
\usepackage{amsmath}
\usepackage{cleveref}
\usepackage{amsfonts}
\usepackage{siunitx}
\usepackage{algorithm}
\usepackage{array}
\usepackage{float}  
\usepackage{microtype}
\usepackage{subcaption}
\usepackage{xcolor}
\usepackage{fmtcount} % for \ordinalnum
\usepackage{multirow}
\usepackage{textcomp}
\usepackage{algpseudocode}
\usepackage{afterpage}
\usepackage{adjustbox}
\usepackage{makecell}
\newcolumntype{C}[1]{>{\centering\arraybackslash}m{#1}}

\newcommand{\xco}[1]{c#1}
\newcommand{\xsi}[1]{s#1}

\begin{document}
\begin{frontmatter}

\title{VR-Caps: A Virtual Environment for Capsule Endoscopy}

\author[a]{Ka\u{g}an \.{I}ncetan}
%\ead{kagan.incetan@boun.edu.tr}

\author[b]{Ibrahim Omer Celik\fnref{fn1}}
%\ead{ibrahim.celik@boun.edu.tr}

\author[a]{Abdulhamid Obeid\fnref{fn1}}
%\ead{abdulhamid.obeid@boun.edu.tr}

\author[a]{Guliz Irem Gokceler}
%\ead{guliz.gokceler@boun.edu.tr}

\author[a]{Kutsev Bengisu Ozyoruk}
%\ead{bengisu.ozyoruk@boun.edu.tr}

\author[c]{Yasin Almalioglu}
%\ead{yasin.almalioglu@cs.ox.ac.uk}

\author[d,g]{Richard J. Chen}
%\ead{richardchen@g.harvard.edu}

\author[d,e,f]{Faisal Mahmood}
%\ead{faisalmahmood@bwh.harvard.edu}

\author[h]{Hunter Gilbert}
%\ead{hbgilbert@lsu.edu}

\author[f]{Nicholas J. Durr}
%\ead{ndurr@jhu.edu}

\author[a]{Mehmet Turan\corref{cor1}}
\ead{mehmet.turan@boun.edu.tr}

\cortext[cor1]{Corresponding Author}

\fntext[fn2]{This work was supported by the Scientific and Technological Research Council of Turkey (TUBITAK) with grant 2232 - The International Fellowship for Outstanding Researchers}
\fntext[fn1]{These authors contributed equally}

\address[a]{Institute of Biomedical Engineering, Bogazici University, Istanbul, Turkey}
\address[b]{ Department of Computer Engineering, Bogazici University, Istanbul, Turkey }
\address[c]{Computer Science Department, University of Oxford, Oxford, UK }
\address[d]{Brigham and Women's Hospital, Harvard Medical School, Boston, MA, USA}
\address[e]{Cancer Data Science, Dana Farber Cancer Institute, Boston, MA, USA}
\address[f]{Cancer Program, Broad Institute of Harvard and MIT, Cambridge, MA, USA}
\address[g]{Department of Biomedical Informatics, Harvard Medical School, Boston, MA, USA}
\address[h]{Deparment of Mechanical and Industrial Engineering, Louisiana State University, Baton Rouge, LA USA}
\address[f]{Department of Biomedical Engineering, Johns Hopkins University (JHU), Baltimore, MD, USA}

\newcommand{\LP}[1]{{\textcolor{blue}{[LP:#1]}}}

\begin{abstract}
Current capsule endoscopes and next-generation robotic capsules for diagnosis and treatment of gastrointestinal diseases are complex cyber-physical platforms that must orchestrate complex software and hardware functions. The desired tasks for these systems include visual localization, depth estimation, 3D mapping, disease detection and segmentation, automated navigation, active control, path realization and optional therapeutic modules such as targeted drug delivery and biopsy sampling. Data-driven algorithms promise to enable many advanced functionalities for capsule endoscopes, but real-world data is challenging to obtain. Physically-realistic simulations providing synthetic data have emerged as a solution to the development of data-driven algorithms. In this work, we present a comprehensive simulation platform for capsule endoscopy operations and introduce VR-Caps, a virtual active capsule environment that simulates a range of normal and abnormal tissue conditions (e.g., inflated, dry, wet etc.) and varied organ types, capsule endoscope designs (e.g., mono, stereo, dual and 360\degree camera), and the type, number, strength, and placement of internal and external magnetic sources that enable active locomotion. VR-Caps makes it possible to both independently or jointly develop, optimize, and test medical imaging and analysis software for the current and next-generation endoscopic capsule systems. To validate this approach, we train state-of-the-art deep neural networks to accomplish various medical image analysis tasks using simulated data from VR-Caps and evaluate the performance of these models on real medical data. Results demonstrate the usefulness and effectiveness of the proposed virtual platform in developing algorithms that quantify fractional coverage, camera trajectory, 3D map reconstruction, and disease classification. All of the code, pre-trained weights and created 3D organ models of the virtual environment with detailed instructions how to setup and use the environment are made publicly available at {\color{blue}\url{https://github.com/CapsuleEndoscope/VirtualCapsuleEndoscopy}} and a video demonstration can be seen in the supplementary videos (Video-I).

\end{abstract}

\end{frontmatter}

\section{Introduction}
Optical colonoscopy is considered to be the gold standard in the early prognosis, diagnosis and intervention of critical upper and lower GI-tract diseases such as colorectal cancer (CRC), Crohn's disease, ulcerative colitis, hemorrhoids or inflammation. Despite colonoscopies demonstrating clinical impact in reducing CRC incidence, the current standard of care for patient screening is invasive and has poor sensitivity in detecting the adenomatous polyps. With an estimated 19 million colonoscopies performed annually in the United States and 6-28\% polyps missed in routine screenings, CRC is the second most prevalent cancer and leading cause of cancer death \citep{lee2017risk}. Due to the fact that small intestines are difficult to access with conventional endoscopes and since patients suffer from heavy pain and discomfort during traditional endoscopy, technologies such as Wireless Capsule Endoscopes (WCEs) have emerged for navigating the entire gastrointestinal tract and identifying adenomatous polyps and other non-polyploid lesions, which would preclude the progression of CRC. 

\begin{figure*}[hbt!]
\centering
    \includegraphics[width=\textwidth]{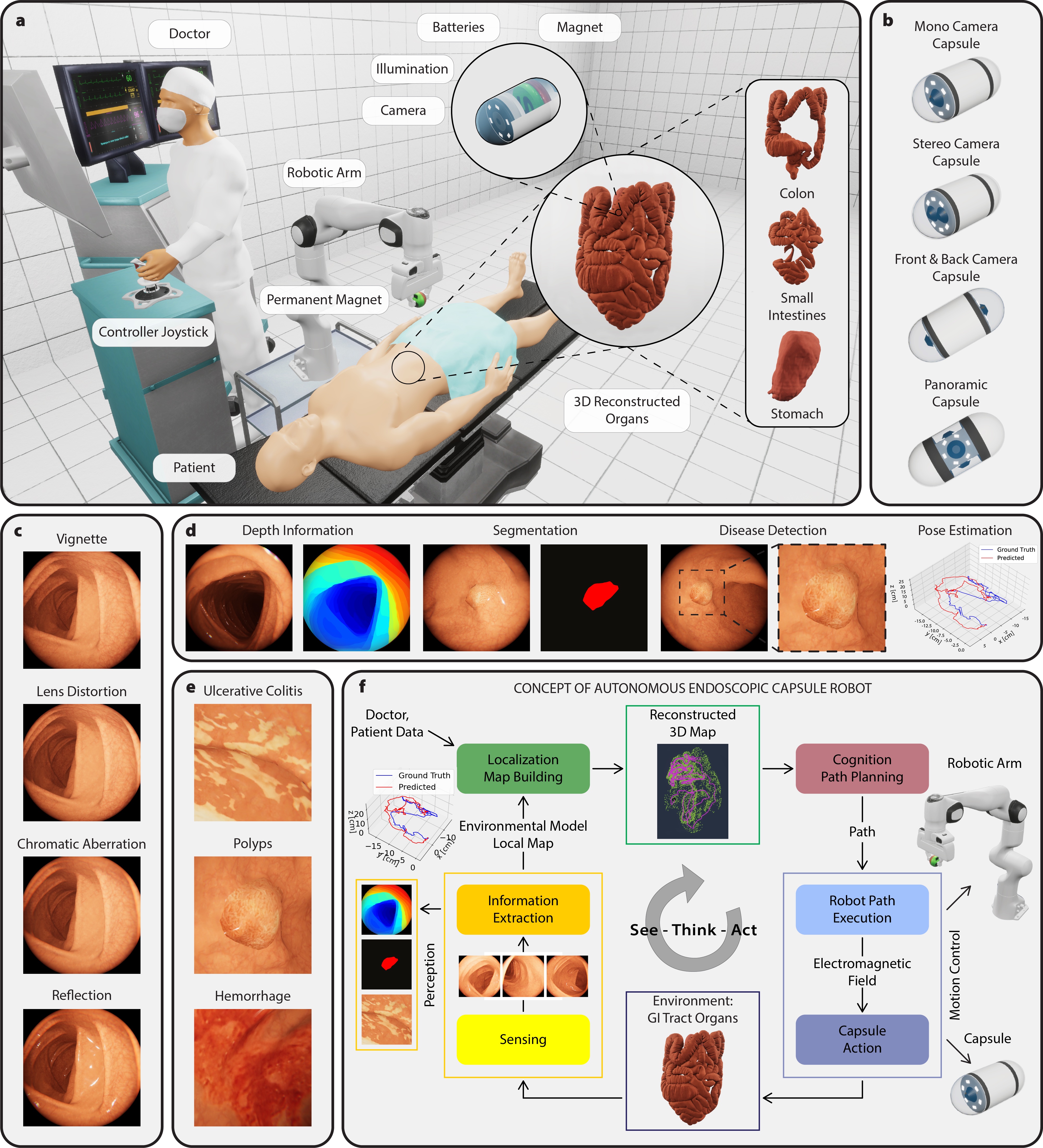}
	\caption{ \textbf{VR-Caps environment.} a) The environment consists of a physician performing the magnetically actuated active capsule endoscopy on a patient by controlling the capsule robot to move throughout the organs with a joystick that is used to move the Franka Emika gripper that holds a permanent magnet. b) Several magnetically actuated capsule designs; mono camera, stereo camera, dual camera (front-back) and a panoramic capsule camera with permanent magnets in different positions and numbers (permanent magnets can also be replaced by electromagnet arrays). c) Post processing effects applied to the capsule camera. d) Some examples of tasks that can be performed in VR-Caps environment: extracting depth information for each frame, disease detection and segmentation, pose estimation from camera frames and generating the capsule robot's trajectory. e) 3 diseases generated in our environment that can be used for disease detection and classification tasks. f) Scheme of the ultimate autonomous capsule robot system.}
    \label{fig:mainfigure}
\end{figure*}

Unlike conventional endoscopes, WCEs are swallowable, pill-like imaging devices that allow for direct visualizations of the GI-tract without requiring bowel preparation or sedation \citep{davies2005colorectal, meining2007effect}. A major limitation of current WCEs is that they progress through the GI system over the course of days. Active locomotion using automated and sophisticated algorithms are needed to guide the capsule robot both to improve screening time and also to provide active functions such as delivery, biopsy and therapeutic interventions \citep{yim2013biopsy, valdastri2012magnetic, pirmoradi2011magnetically, yim2011design}. If a lesion is detected, an automated path planning and execution functionality should reach the lesion in the most effective and shortest way. Once the lesion is reached, therapeutic actions such as drug release or biopsy can follow.

To achieve highly adaptive and patient specific control, localization, mapping, path planning and path realization mechanisms, deep learning will be a key enabler, having previously demonstrated learning and domain adaptation capability in many computer vision and robotic control fields \cite{mahmood2019deepadversarial}. One of the biggest disadvantages of deep learning techniques is the fact that large networks need massive amounts of domain-specific data for training. Research in recent years has shown that large amounts of synthetic data can improve the performance of learning-based robotic control and vision algorithms and can ameliorate the difficulty and expense of obtaining real measurements in a variety of contexts \citep{peng2015learning,hattori2015learning,su2015render,qiu2016unrealcv}. Specifically in the context of active capsule robots, previous works have simulated some aspects successfully. \cite{mura2016vision} developed a model-based virtual haptic feedback mechanism to assist the navigation of endoscopic capsule robots. More recently, \citep{abu2019proposed} proposed a simulation environment particularly for the locomotion mechanism of endoscopic capsule robots. 

Some simulations have targeted vision-related tasks for endoscopy. \cite{mahmood2018unsupervised,mahmood2018deepspie,mahmood2018deepee} proposed an approach to generate synthetic endoscopy images through deep adversarial training and demonstrated its success in depth estimation task. \cite{he2018hookworm} explored automatic hookworm detection from WCE using a CNN to classify frames in video sequences. \cite{turan2018deep} developed one of the first WCEs with visual odometry in the GI using RCNNs, and later developed a deep reinforcement learning-based approach that is able to learn a policy for navigation in real porcine tissue. Though deep learning has potential in automating polyp detection via WCEs, these systems require enormous amounts of data to generalize to unseen patient tissue and organ topology. Cross-patient network adaptation is another well-known challenge in endoscopy, as low-level patient-specific texture details such as vascular patterns can obscure important clinically-relevant details that should be generalized across patients. As a result, the potential benefits of deep learning have failed to translate in the technological and clinical deployment of autonomous WCEs. 

Even though, prior simulation approaches have been limited to only a one part of the overall WCE system, in this work, we address the lack of a realistic virtual test-bed platform for WCEs by introducing a new simulation environment, VR-Caps. Using advanced rendering techniques, VR-Caps can generate fully-labeled and realistic synthetic data that is consistent with the topology and texture of real organs for control, navigation and machine vision related tasks. Supported by the fact that deep learning has already proven its domain adaptation capability in various fields of medical image analysis and device control fields \citep{mahmood2018unsupervised,mahmood2018deepee,mura2016vision}, VR-Caps can facilitate significant improvements in medical imaging and device control applications. In the context of both active and passive capsule endoscopy, VR-Caps offers numerous advantages over physical testing:

\begin{itemize}
 \item VR-Caps can accelerate the design, testing and optimization process for software and related hardware components that directly affect medical image analysis applications such as number and placement of cameras, camera specifications, control accuracy of the robot etc.;
 \item The marginal cost of synthetic data is low compared to real data;
 \item System properties and parameter values can be easily altered to assess sensitivity and robustness;
 \item VR-Caps carries no risks to live animals or humans;
 \item VR-Caps can offer reproducibility, which is valuable in the scientific pursuit of new algorithms;
 \item The prevalence of rare diseases can be exaggerated in VR-Caps to provide data that may be infeasible or impossible to obtain from human study participants.
\end{itemize}

VR-Caps facilitates changes in the number, strength, and placement of magnetic sources; organ type (stomach, small intestine and colon); conditions (e.g., inflated, dry, wet etc.); deformation and peristaltic motion; camera number and locations (e.g., mono, stereo, dual and 360\degree); types of illumination (e.g., point light source, LED arrays). It also facilitates inter-patient diversity (e.g., availability of CT scans and corresponding 3D organ models from different patients) with different abnormalities (e.g., varying size of tumors and different diseases with various grades). 

The paper is organized as follows: \Cref{sec:setup} describes our simulation environment and gives detailed information about the pipeline of generating 3D models and realistic organ textures. Properties required to simulate physical interactions of capsule endoscopy are also explained in this section. In \Cref{sec:eva_tasks_res}, we demonstrate 6 use-cases of the environment and the achieved results. Finally, \Cref{sec:discussionandconclusion} presents the concluding remarks and future works.

\begin{figure*}
\centering
    \includegraphics[width=\textwidth]{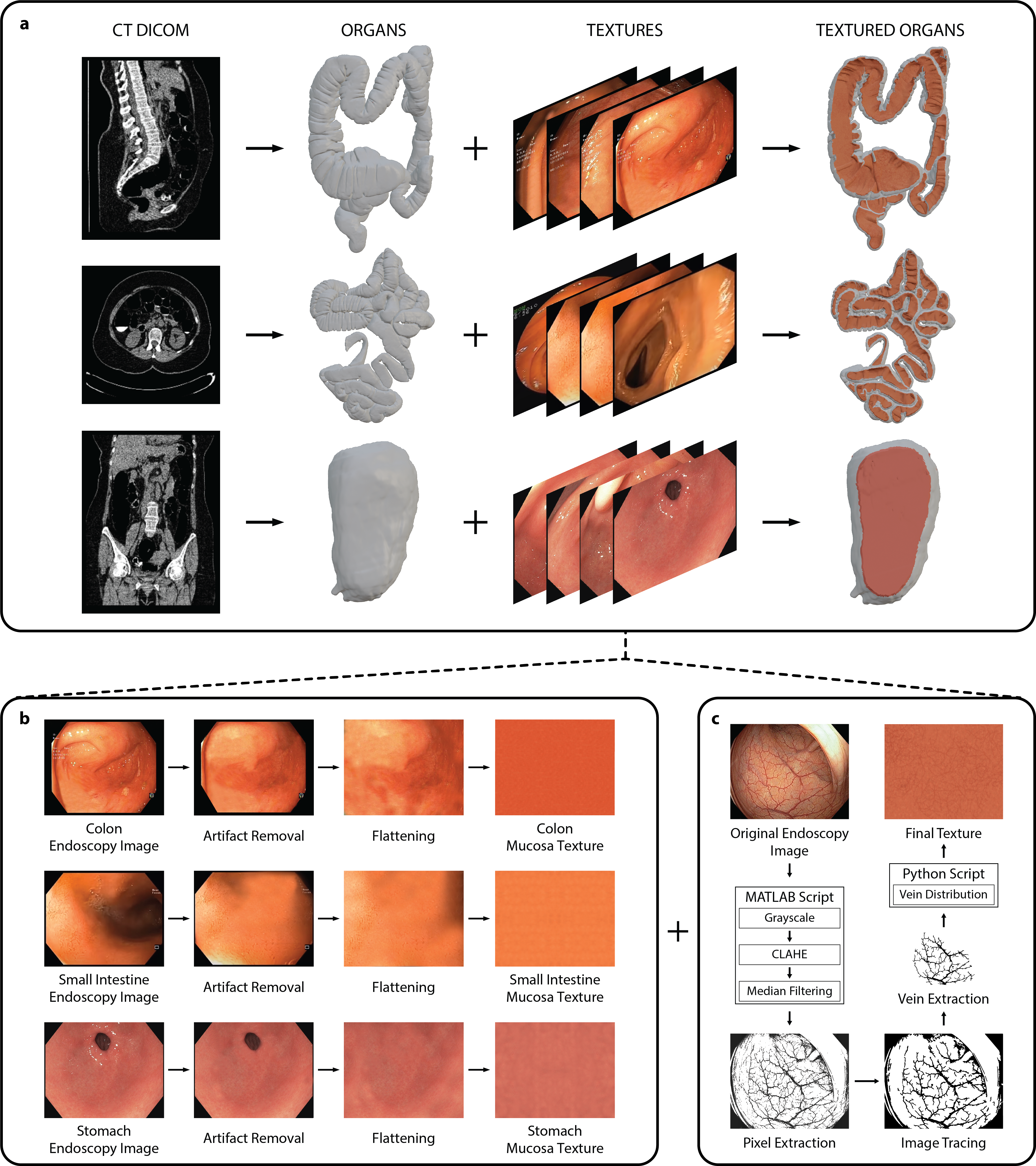}
    \caption{\textbf{3D organ construction and texture assignment process.} a) We use 3D models reconstructed from real patient CT scans and create textures based on analysis of real endoscopic videos and assign the generated textures to the 3D models by using UV mapping technique.
    b) The flow of creating the mucosa textures from real endoscopy images.The flow of creating the mucosa textures from real endoscopy images. We first remove artifacts and reflections on the endoscopy image and then select a uniform region in terms of color and expand it to create the main mucosa texture
    c) The pipeline for adding veins to the mucosa texture. We extract veins from real endoscopy images using MATLAB and assign them on a texture image using a Gaussian distribution that forms relevant vein network.}
    \label{fig:modelgeneration}
\end{figure*}

\section{VR-Caps Environment}
\label{sec:setup}
The VR-Caps environment is built using the popular graphics engine and a real-time 3D development platform  Unity with the integration of Simulation Open Framework Architecture (SOFA) \citep{faure2012sofa} and Unity Machine Learning Agents Toolkit (ML-Agents) \citep{juliani2018unity}. The SOFA is an open source framework primarily targeted at medical simulation research that allows to create complex medical simulations such as organ deformation and collisions. The ML-Agents is an open-source plug-in dedicated for machine learning applications to be integrated into Unity platform. It enables Unity to serve as an interactive platform for the training of neural networks based intelligent agents using state-of-the art approaches in deep reinforcement learning and imitation learning. 

\subsection{Generating 3D Organ Models}

To represent the 3D geometry of GI organs accurately, we use computed tomography (CT) images in DICOM format, which are publicly available on The Cancer Imaging Archive (TCIA) \citep{clark2013cancer}. As the archive includes various sets for different parts of the body, for colon and small intestine, we use \citep{COLONDATAsmith2015data} consisting of 825 subjects and 347 of them are differentiated into three classes: 243 patients with no polyps, 69 patients with 6 to 9 mm sized polyps, and 35 patients with polyps larger than 10 mm and for stomach, we use \citep{STOMACHDATAlucchesi10radiology} consisting of 46 subjects. Although the dataset includes two subsets in terms of patient position (supine and prone position) during imaging, the supine position sets is used because this is the position of the patient during capsule endoscopy. The dataset also has different models of the organs with and without the cancerous lumps and we use them to have real polyps in the reconstructed models.

Using an open-source medical image reconstruction software (i.e., InVesalius), 3D organ models are created from CT scans. The reconstructed 3D model is then imported into Blender for further processing. As shown in \Cref{fig:modelgeneration}-a, the imported model consists of bones, fat, skin, and other artifacts that are removed so that only the geometries of the colon, small intestines and stomach remain. As a next step, textures are created using the Kvasir \citep{pogorelov2017kvasir} that consists of real endoscopy images taken from different patients classified according to the different parts of the GI tract. In order to create the main mucosa texture from the Kvasir dataset, various endoscopy images are stitched together and applied on the model inner surface to generate clear, non-blurry and continuous mucosa walls (see \Cref{fig:modelgeneration}-b). After generating the main mucosa textures, veins are applied on the walls by extracting vein networks from Kvasir dataset images. To extract the veins from real endoscopic images, we convert images into gray-scale and perform median filtering on the pixels. A contrast-limited adaptive histogram equalization is used to enhance the contrast of the gray-scale image and to make the veins darker than its surroundings as shown in \Cref{fig:modelgeneration}-c. Extracted veins are applied on the mucosa texture images, using random distributions for location, rotation and size of the veins with empirically determined mean and standard deviation values. In addition to normal tissue, the textures of the disease regions are also created from Kvasir videos that contain instances of the corresponding diseases. \Cref{fig:mainfigure}-e shows frames with diseases instances, while \Cref{fig:mainfigure}-c shows healthy regions of the organs.
Finally, we unwrap the 3D model and the mesh model is divided into rectangular segments that are projected on the created UV texture map uniformly in a repetitive manner throughout the model.

\begin{figure}[hbt!]
\centering
    \includegraphics[width=\columnwidth]{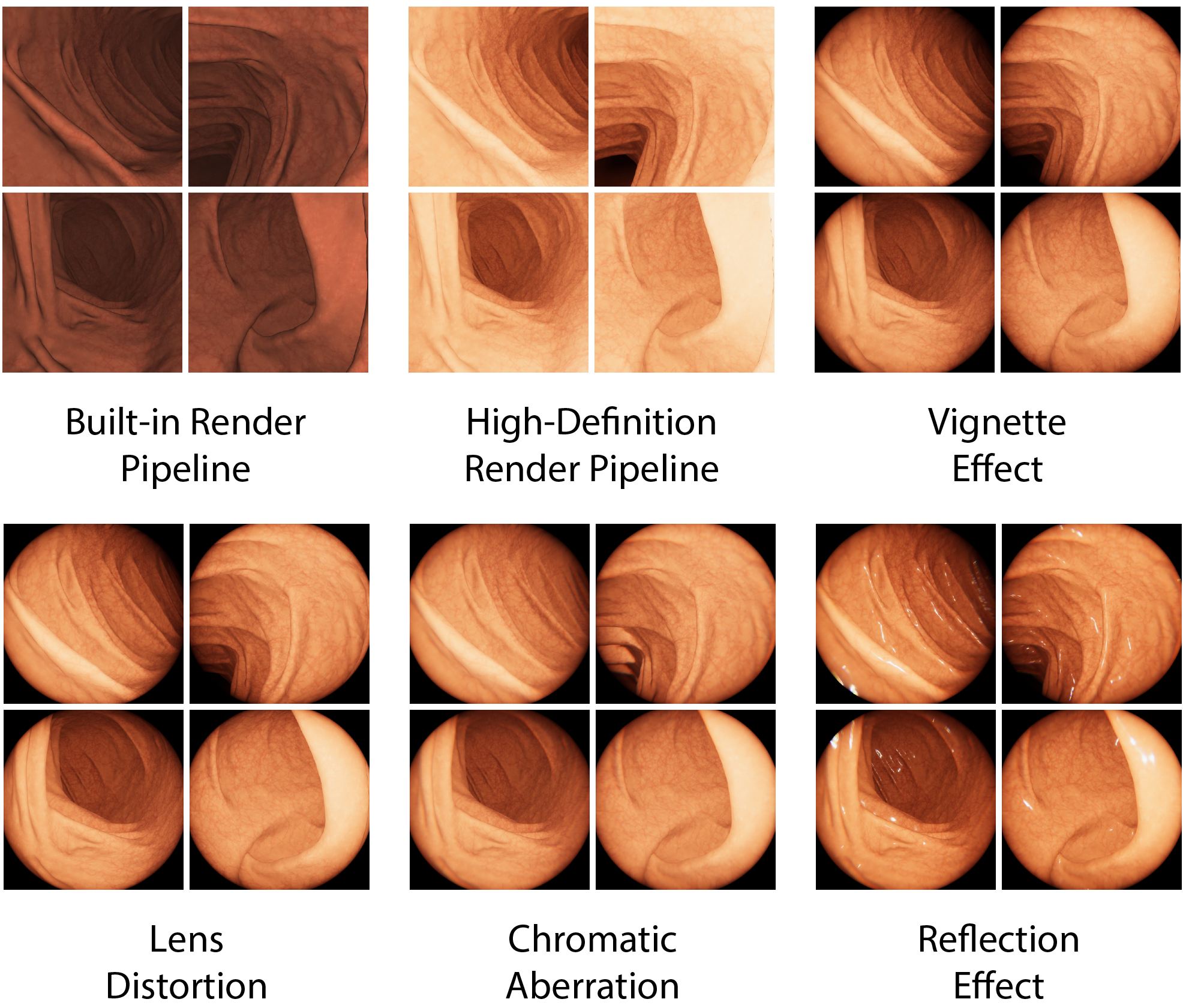}
    \caption{\textbf{Post-processing pipeline.} In order to generate virtual endoscopy images that can be similar to real endoscopic images, we improve the rendering quality of the simulation by integrating the HDRP. This pipeline provides the tools to apply some post-processing effects on the camera with adjustable parameters; such as vignette, lens distortion, and chromatic aberration. Lastly, we introduce a wet surface reflection property that interacts with the light source.}
    \label{fig:pipeline}
\end{figure}

\subsection{Rendering Tool}

As the default pipeline of Unity does not provide high quality and realistic visuals, we integrate the High Definition Render Pipeline (HDRP) which mainly works on differentiating materials in different lighting conditions while unifying illumination so that all objects in the scene receive and interact with light in the same manner. 

The HDRP shaders provide several options that facilitate obtaining more realistic visuals and imitating real endoscopy images. For instance, adding white coat mask to the organ material creates reflection effect in 3D organ surfaces when hit by any light source. Endoscopic camera views are also mimicked using HDRP such as vignetting, fish-eye distortion, and chromatic aberration. Vignetting is the peripheral darkening of the endoscopy image, and chromatic aberration which usually occurs when the lens fails to converge all colour wavelengths to the same point on the focal plane causes the edges between high contrast areas to appear blurred. These effects are applied directly to the image buffer of the virtual camera which provides real time rendering opportunities. The enhanced image quality using the HDRP as well the post-processing effects can be seen in \Cref{fig:pipeline}.

\begin{figure}[hbt!]
\centering
    \includegraphics[width=\columnwidth]{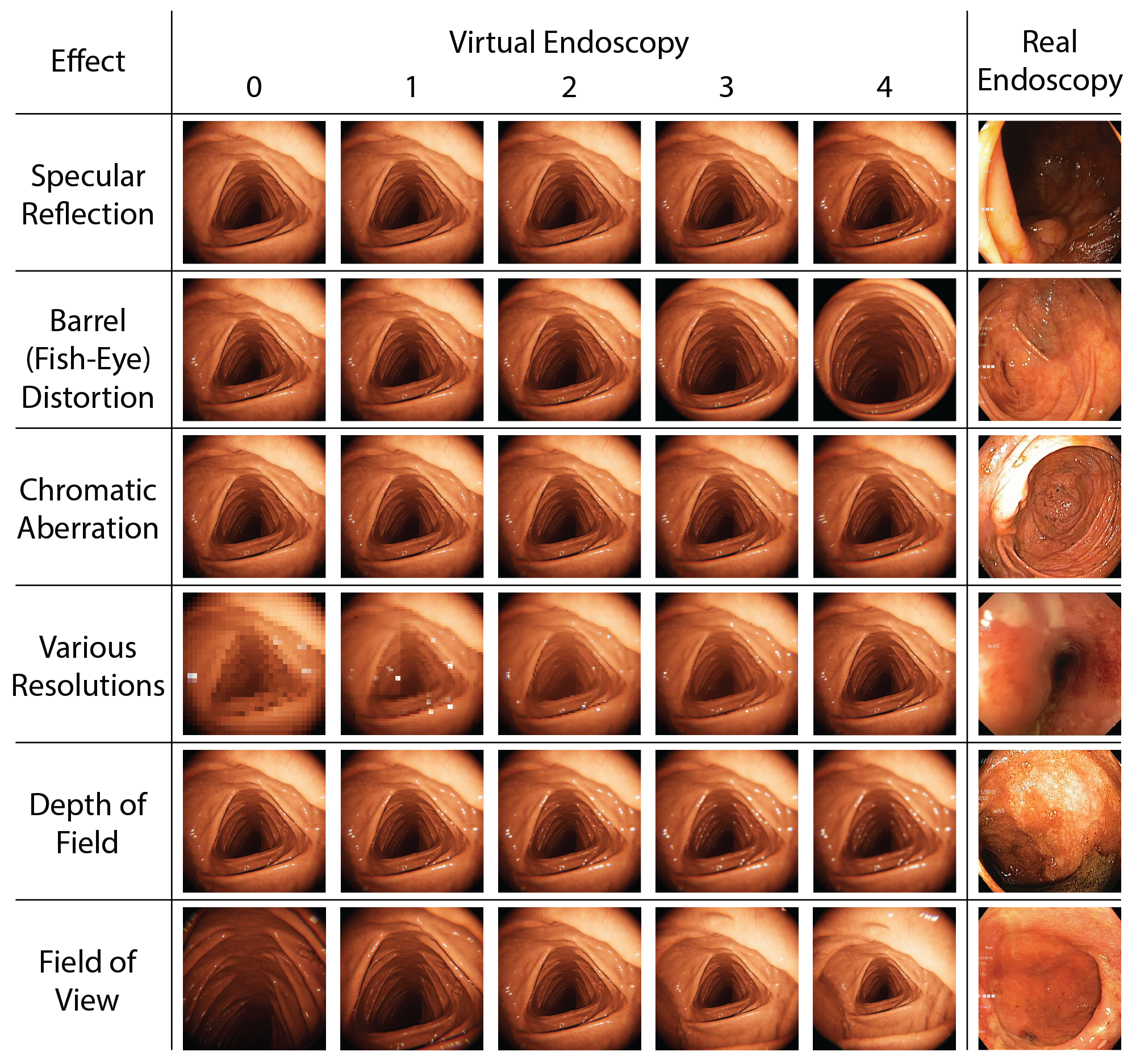}
	\caption{\textbf{Adjustable effects to mimic real endoscopy images in VR-Caps} Real endoscopy images have some specific artefacts and imperfections resulting from camera features and lighting conditions. In VR-Caps environment, we mimic some of these effects, namely: specular reflection, barrel distortion, chromatic aberration, various resolutions, depth of field and field of view. In this figure, we show the possibility of adjusting intensity and occurrences of those effects for 5 different levels. Here, grade 0 to 4 denotes the increasing level of intensity for each effect.}
    \label{fig:effects}
\end{figure}

To show that our environment facilitates changes of aforementioned effects with some other variations such as different camera resolution, depth of field and field of view, we provide \Cref{fig:effects} in which the different level intensity of specularities and other post-processing effects are applied from level 0-to-4 where grade 0 indicates the lowest possible intensity for each effect. 
Besides these camera effects and artifacts, real endoscopic videos contain a large amount of liquid effects seen in the GI organ walls. Together with the pumped air during endoscopic operation, this creates bubbles of varying sizes and strong mirror reflections and specularities in the recorded videos. Also, partly-digested food and gastric fluid cause several floating particles known as yellow or white chyme appearing at different locations inside the GI. These particles can distract the motion of the capsule endoscope and occlude the camera lens. In VR-Caps, we mimic these occurrences by creating virtual bubbles and chyme using both individual and bounded particle systems with variable fluid effects on the organ walls. Bubble particles are modeled as bouncing spheres of different sizes, different intensities of random motions. They tend to rise above the surface as in the real endoscopic videos (i.e., in the opposite direction to the gravity). Floating particles are designed as separate objects of varying sizes, and they move along the way to the end of the colon. These effects can be seen in \Cref{fig:bubbleChyme} and in the supplementary videos (Video-II).

\subsection{Deformation}

Due to its physical accuracy, the FEM (Finite Element Analysis), which is an approximate discrete representation of soft tissue that is obtained by dividing the volume of tissue into smaller elements, is widely used in bio-mechanical modeling of soft tissues in various medical fields such as image-guided hepatic surgery, computer-integrated neurosurgery, whole-body medical image registration, and surgical simulations \citep{zhang2017deformable}. In VR-Caps, in order to simulate the soft deformable mechanics of organs, we integrate SofaAPAPI-Unity3D which is an interface that enables Unity's PhysX Engine to leverage SOFA's more physically accurate models for tissue deformation.

The basic equation solved to model the tissue deformation is the conservation of momentum:

\begin{equation}
\rho \frac{d\vec{v}}{dt} = \rho \vec{b} + \nabla \cdot \sigma
\end{equation}
where $\rho$ is the mass density of the tissue, $\vec{b}$ is the density of external forces per unit mass, and $\sigma$ is the Cauchy stress tensor, which is related to the deformation of the tissue by a suitable constitutive law. SOFA uses a linear elastic (Hookean) material model and it solves the equation in the weak form after the inclusion of boundary conditions that constrain the organ. 

As the finite element method is computationally expensive, we also integrate a deformation model which is intended only to reproduce a deflection of similar appearance to the physically accurate one. The elastic deformation which occurs when the capsule robot is in contact with the organ walls is simulated by the Collider component which is offered in Unity's default physics engine and used for defining the shape of the object for the physical collisions. When there is a contact between the capsule robot and the organ, the Collider component stores the position where the collision occurs and the force that capsule robot applies to the organ. The displacement amount of vertices of the organ mesh under the force is set so that it gets its maximum value at the contact point and reduces as the distance between the contact point and position of adjacent vertices increases. An attenuated force at each vertex, $\vec{F}_v$ is calculated by using the inverse-square law;

\begin{equation}
\vec{F}_v = \dfrac{\vec{F}}{d^{2} + 1},
\end{equation}
where the original force $\vec{F}$ is divided by the distance squared added 1, to achieve a full strength force at the distance close to zero and avoid the force going to infinity. With a given attenuation force, the change in mesh displacement in unit time interval is defined as:

\begin{equation}
\Delta{\vec{v}} = \dfrac{\vec{F}}{m}\Delta{t}.
\end{equation}

For simplicity, we assumed the mass, m, as 1 for each vertex, and end up with,

\begin{equation}
\Delta{\vec{v}} = \vec{F}\Delta{t}.
\end{equation}

So that the position of vertices, $\vec{x}$, are updated by following adjustment;

\begin{equation}
\Delta{\vec{x}} = \vec{v}\Delta{t}.
\end{equation}

As far as the force is applied, vertices keep moving without any limitations, which is not realistic due to the elasticity properties of objects. So we introduce a spring-like force controlled by an adjustable factor that returns the vertices to their original position gradually. The spring force causes some oscillations which are also prevented by adding a damping effect with resistance constant $\mu$, it is a factor that decreases velocity over time,

\begin{equation}
\vec{v_d} = \vec{v}(1-\mu \Delta{t}).
\end{equation}

\begin{figure}[hbt!]
\centering
    \includegraphics[width=\columnwidth]{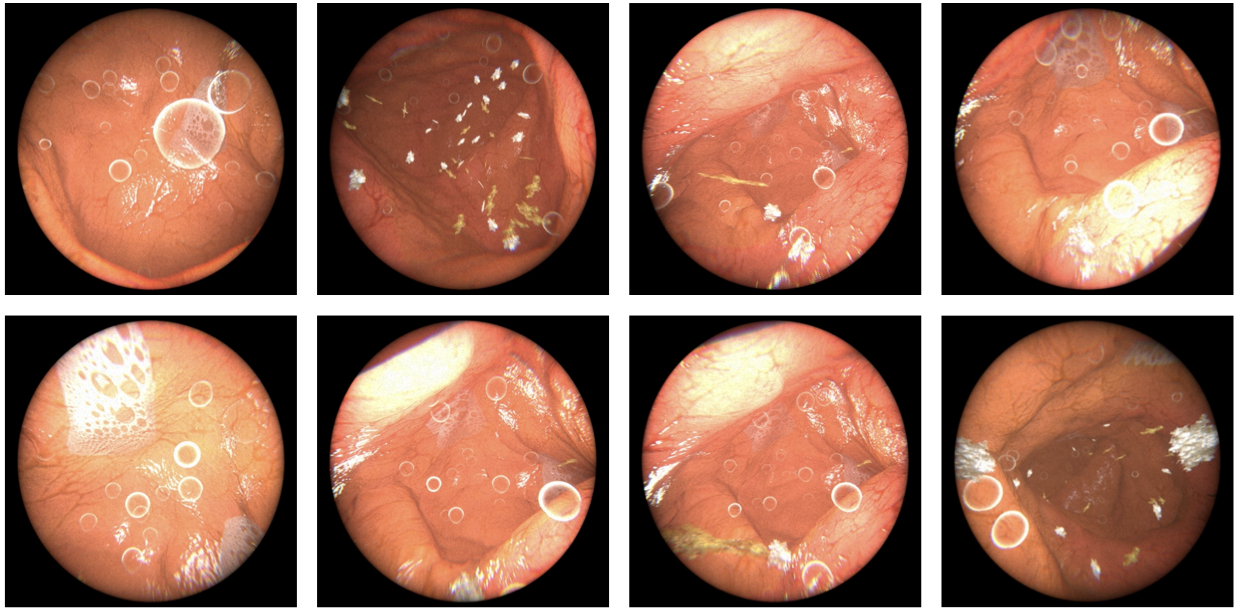}
    \caption{\textbf{Bubbles and chymes} In the VR-Caps platform, we mimic the bubbles, chyme and reflection effects caused by liquids inside the GI tract and randomly floating particles (e.g., chyme) that are party-digested food appearing in yellow and white colour. By creating particle system objects in different sizes and configurations, both bubbles and chyme are simulated in VR-Caps.}
    \label{fig:bubbleChyme}
\end{figure}
\subsection{Peristaltic Motion}
Peristalsis is the involuntary motion in the GI-tract that occurs in a wavelike motion progressively through the length of the organs pushing down the food until excretion. During the capsule endoscopy session the patient should be in a fasting state for almost 12 hours. The migrating motor complex is the motility process that happens during the fasting (interdigestive) period. It is an electromechanical activity that occurs in a regular cycle every 1,5 to 2 hours that consists of four phases that lasts from 85 to 115 minutes. The first phase is an inactive period without contractions that lasts for 40-60\% of the cycle length. The second phase consists of increasingly strong contractions that lasts 20-30\% of the cycle length. The third phase consists of strong contractions that lasts 5-10 minutes. The fourth phase is a recovery phase in which the organs return into the inactive first phase to repeat the cycle \citep{MMC1, feher20178}. The contractions exhibit variety in different regions of the organ which cause different propagation velocities throughout the GI-path. According to the works of \cite{MMCV1, MMCV2}, these velocities and duration of cycle phases highly depend on the genetic background and vary across individuals.

The small intestine and colon consist of bends and curves that can be used to segment the length into discrete segments. Since the capsule robot move in different direction for each segment, we can describe the peristalsis motion way by subtracting initial and final position of each segment:

\begin{equation}
\vec{v} = \dfrac{P_e - P_s}{\lVert P_e - P_s \rVert},
\end{equation}
where $P_e$ and $P_s$ represent end and start point of organ segments, respectively. In the Unity environment, the velocity (cm per minute) under the peristaltic motion and movement direction of capsule robot are simulated based on its residing segment type. Finally, the force faced by the capsule robot in unit time interval due to peristalsis is defined by:
 
\begin{equation}
\vec{F}_p = \beta(\vec{v_p} - \vec{u}),
\end{equation}
where $\vec{v_p}=\alpha \vec{v}$ is the propagation velocity which is a scaled multiple of the propagation direction, and ${\vec{u}}$ is the current velocity of the capsule robot.

To achieve the physical movement of the contractions in our simulation environment, meshes of the 3D organ in question is moved in a wave-like motion. To achieve that, each vertex is moved by a sine wave which gives the motion of contraction and expansion of the organ as the capsule robot propagates through. The Collider component attached to the deformed organ is also updated with its mesh to prompt interaction of the capsule robot with the contractions. Due to all the variability in the duration of the phases and the propagation velocities, the peristalsis motion algorithm is designed in a way that user can adjust the duration and velocity parameters.

\subsection{Frictional Resistance}
The friction is not completely modeled by the Coulomb friction law due to viscoelastic behaviour of the intestine \citep{accoto2001measurements, friction2008}. Various frictional forces are applied during the navigation of the capsule robot and these forces are a function of the robot velocity $\vec{v}$, generally acting along the unit vector opposite to the velocity, $\vec{d}_f = -\vec{v}/\lVert \vec{v}\lVert$ In general, total frictional force $\vec{f}(\vec{v})$ is modeled as follows: 

\begin{equation}
\label{eq:c_friction}
\begin{split}
\vec{f} = {\vec{f_c}} + {\vec{f_e}} + {\vec{f_v}}
\end{split}
\end{equation} where $\vec{f}_c  = -\mu \lVert N \lVert \vec{d}_f$ is the Coulomb friction, which is a function of the normal contact force $N$,  ${f}_e$ is the environmental resistance caused by hoop stress and elastic deformation of intestinal wall, and ${f}_v$ is the visco-adhesive force which occurs due to intestinal mucus \citep{friction}. The environmental resistance can be expressed as: 

\begin{equation}
\label{eq:e_friction}
\begin{split}
\vec{{f}_e} = P(x) \vec{S}\sin(\theta)
\end{split}
\end{equation}
where $P(x)$ is the pressure applied to contact surface area normal $\vec{S}$ (generally along the axis of the lumen as a result of radial symmetry and contact around the capsule) and the $\theta$ is the angle of the skew force. This force results from the viscoelasticity of the tissue, which causes the hoop stress at the front of the capsule, where the lumen is stretching open, to be greater than the hoop stress at the back of the capsule, \begin{figure}[t]
\centering
    \includegraphics[width=\columnwidth]{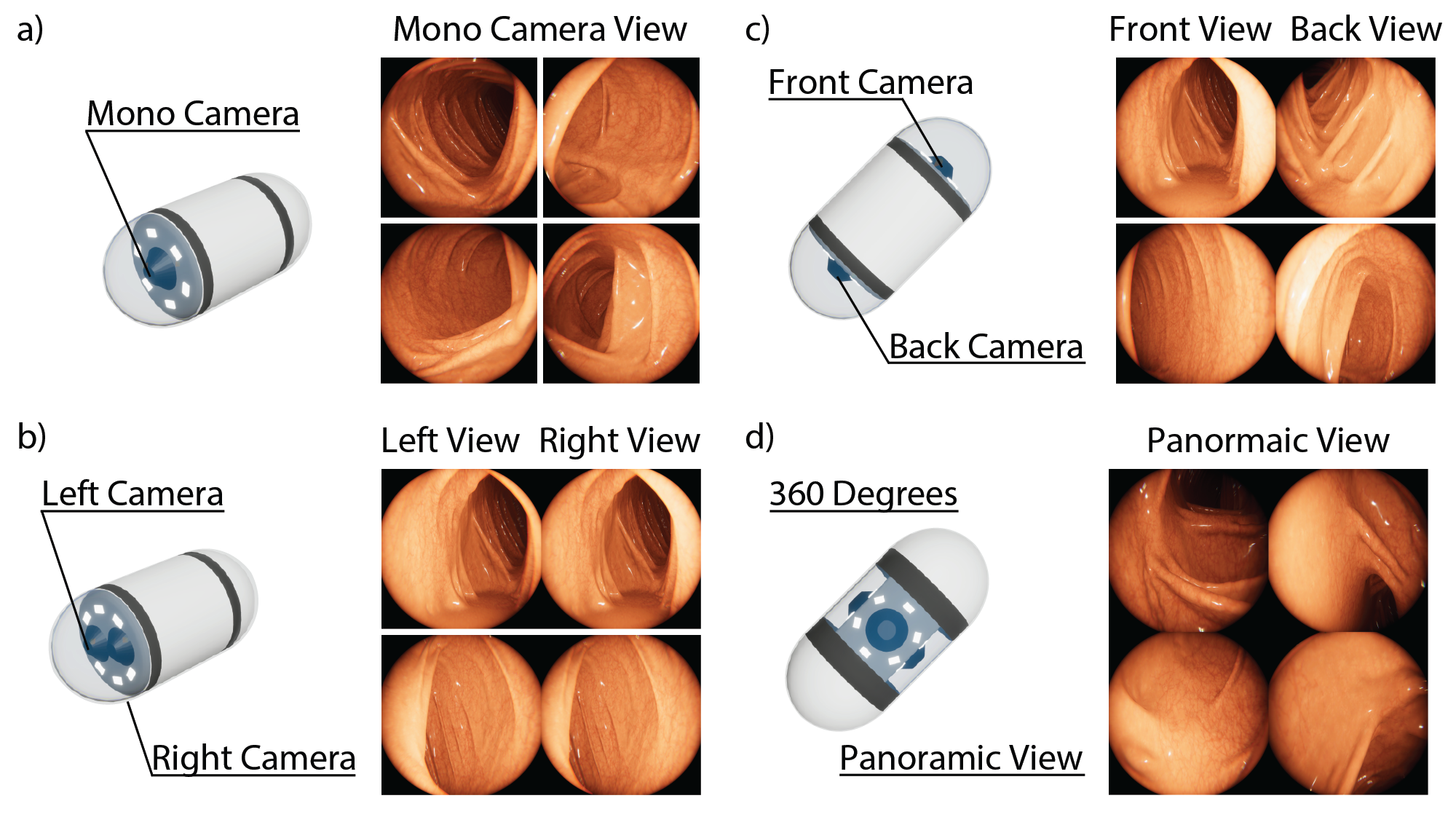}
    \caption{\textbf{Capsule designs with various camera configurations.} We have designed and employed four endoscopic capsule robots with different number of cameras in different orientations. The designs and the corresponding output frames from all camera views are shown in the figure. a) The regular capsule which contains a single camera. b) The stereo camera configuration that contains two cameras on one end of the capsule with an offset between them. c) Dual camera design that consists of a camera at front-end and a second camera at back-end of the capsule endoscope. d) A panoramic design consisting of four cameras placed on the sides of capsule robot body with 90\degree  rotational offset to each other, resulting in a 360\degree view. A video showing the recordings using these different configurations can be seen in the supplementary videos (Video-III).
    } 
    \label{fig:capsules}
\end{figure} where the lumen has already been stretched. Visco-adhesive resistance is directly correlated with the velocity of the capsule robot and it is defined as: 

\begin{equation}
\label{eq:e_friction}
\begin{split}
\vec{{f_v}} = -\gamma \vec{v}
\end{split}
\end{equation} where $\gamma$ is the viscosity coefficient of intestinal mucus. We include all three types of frictional forces in our simulation by relying on the measurements provided in \citep{friction}. Since the study has shown a logarithmic correlation between total frictional force and capsule velocity we define the following partial function to simulate the effect of total resistance:

\begin{equation}
{f} =  \begin{cases} 
     C & \lVert \vec{v}\lVert\leq 1 \\
      a\log(b\lVert \vec{v}\lVert +c) + C & \lVert \vec{v}\lVert > 1
  \end{cases}
\end{equation}
To estimate parameters, we fit a curve to the published data using a non-linear least-squares optimization method. Estimated parameters are $a= 55.04$,  $b= 0.23$, $c= 1.04$, $C=100$. 

\subsection{Magnetic Field}
Magnets are commonly used as external locomotion devices for capsule endoscopes \citep{ciuti2011capsule}. Therefore, in VR-Caps, we provide option of using a cylinder permanent magnet held by a 7 DOF robotic arm as a locomotion device that interacts with another permanent magnet inside the capsule robot. Although Unity engine has built-in physics provided by Nvidia PhysX for most of the physical interactions between objects, there is no existing built-in model for simulating magnetic field \citep{juliani2018unity}. Therefore, a 3rd party model \cite{magnetodynamics} is integrated into VR-Caps which models dipole-dipole interactions. In the model, a dipole’s contribution to the magnetic field at a point is calculated as follows:

\begin{equation}
\label{eq:magneto}
\begin{split} 
    \vec{B}_{\vec{m}}(\vec{r}) = \dfrac{\mu_{0}}{4\pi}\dfrac{1}{r^{3}}\left[3(\vec{m}\cdot\hat{r})\hat{r} - \vec{m}\right]
\end{split}
\end{equation} where  $\mu_{0}$ is the permeability of free space ($4\pi \times 10^{-7} \dfrac{N}{A^{2}}$), $\vec{r}$ is the displacement vector between the dipole and the target point, $r$ is the magnitude of the vector and $\hat{r}$ is its normalized form, and $\vec{m}$ is the dipole vector. The torque on a dipole due to total field $\vec{B}$ at a point is: 

\begin{equation}
\label{eq:magnetoTorque}
\begin{split}
    \vec{N}= \vec{m} \times \vec{B}
\end{split}
\end{equation}
The force on a dipole due to the same field is:

\begin{equation}
\label{eq:magnetoForce}
\begin{split}
    \vec{F}= \nabla \left( \vec{m} \cdot \vec{B} \right) = \vec{m} \times \left ( \nabla \times \vec{B} \right) + \left(\vec{m} \cdot \nabla \right) \vec{B}
\end{split}
\end{equation}

\begin{figure*}[hbt!]
\centering
    \includegraphics[width=\textwidth]{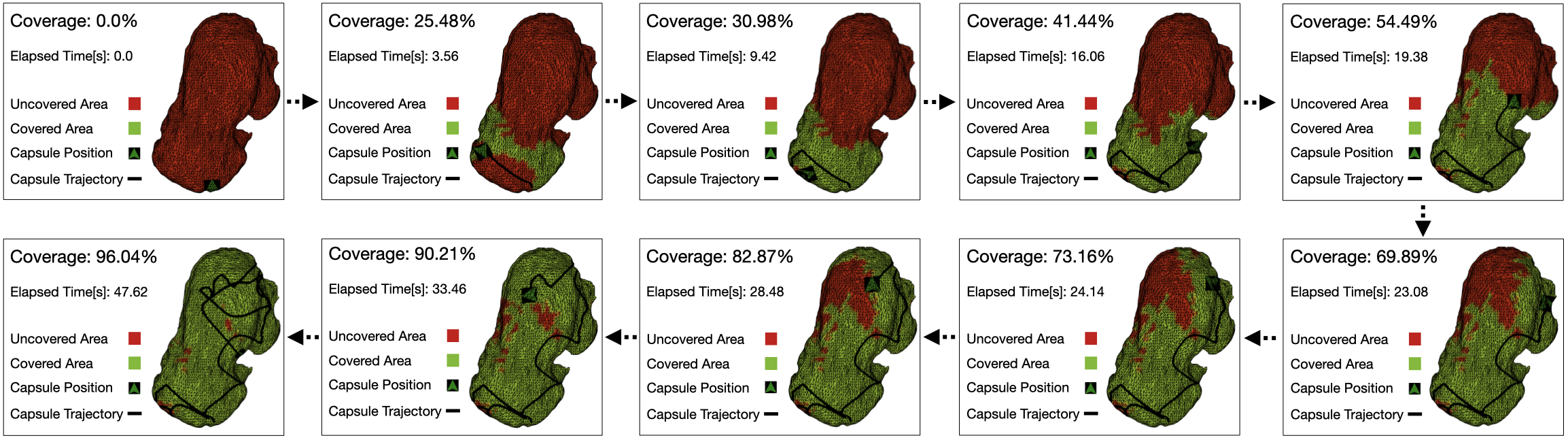}
	\caption{\textbf{Area coverage results} A Deep Reinforcement Learning (DRL) based active control method is developed for the autonomous monitoring of the organ (stomach in this case). The capsule robot is trained with the target of maximizing the coverage of the stomach in a minimum operation time. The coverage is defined as the number of vertices seen by the capsule camera divided by the total number of vertices. The reward is the linear increase in coverage for each time $t$ as seen in \Cref{eq:reward}. This figure visually illustrates the results for the coverage task at 10 time instances during the inference process. Stomach is represented as 3D point cloud projected onto a 2D plane, and green and red regions depict covered and uncovered parts of the stomach respectively. A coverage rate of  96.04\% was achieved in 47.62 seconds.}
    \label{fig:coverage_task}
\end{figure*}

\subsection{Robotic Arm}
As a holder for the permanent magnet, the exact and rigged 3D model of Franka Emika's 7 DOF robotic arm is used in the simulation environment. In order to mimic real capabilities as well as constraints of the robot, inverse kinematics is applied by following a similar approach in \citep{nammoto2012analytical} and joint angles for the desired magnet locations are computed. The transformation matrix of the end-point relative to the base of the robot is defined by a recursive matrix-vector operation in the form of

\begin{equation}
    \label{eq:endpoint}
    \begin{split}
        \begin{bmatrix}
        R^{n}_{0} & T^{n}_{0} \\
        0 \cdots 0 & 1
        \end{bmatrix}= {A_{1}^{0}}(\theta_{0}) {A_{2}^{1}}(\theta_{1}) \cdots {A_{i}^{i-1}}(\theta_{i})
    \end{split}
\end{equation}
where  ${A_{i}^{i-1}}(\theta_{i})$ is homogeneous transformation matrix with Denavit-Hertenberg parameters at a given configuration from the coordinate system of the (i-1)th joint to the ith joint and it is given as

\begin{equation}
\label{eq:inversekin}
{
A_{i-1}^{i}(\theta_{i})=
   \begin{bmatrix}
     \xco{\theta_{i}}&\xsi{\theta_{i}}&0&-a_{i} \\
     -\xsi{\theta_{i}}\xco{\alpha_{i}}&\xco{\theta_{i}}\xco{\alpha_{i}}&\xsi{\alpha_{i}}&-d_{i}\xsi{\alpha_{i}} \\
     \xsi{\theta_{i}}\xsi{\alpha_{i}}&-\xco{\theta_{i}}\xsi{\alpha_{i}}& \xco{\alpha_{i}}&-d_{i}\xco{\alpha_{i}} \\
     0&0&0&1
   \end{bmatrix}}
\end{equation}

\subsection{Endoscopic Capsule Camera Parameters}
Simulating a realistic endoscopic capsule camera has a considerable impact on generating realistically looking images. For this reason, we calibrate two commercially available capsule endoscope cameras, Mirocam and Pillcam, using the pinhole camera model with non-linear radial lens distortions given by Camera Calibrator App from \cite{camera_calibration}. As per model, the standard affine camera model is used including a focal length $(f_x, f_y)$, principal points offset $(c_x, c_y)$, and lens shifting coefficient $s$ and these parameter sets are provided on the Github page. By adjusting these parameter settings, any type of capsule endoscope cameras can be mimicked in the simulation environment. In addition to intrinsic parameters of endoscopic cameras, we also provide various camera configurations in terms of the number and placement of the integrated cameras. Mono camera design, stereo, dual and 360\degree camera designs can be seen in \Cref{fig:capsules} and accessed in the Github source code.

\section{Evaluated Tasks and Results}
\label{sec:eva_tasks_res}

\begin{figure*}
\centering
    \includegraphics[width=\textwidth]{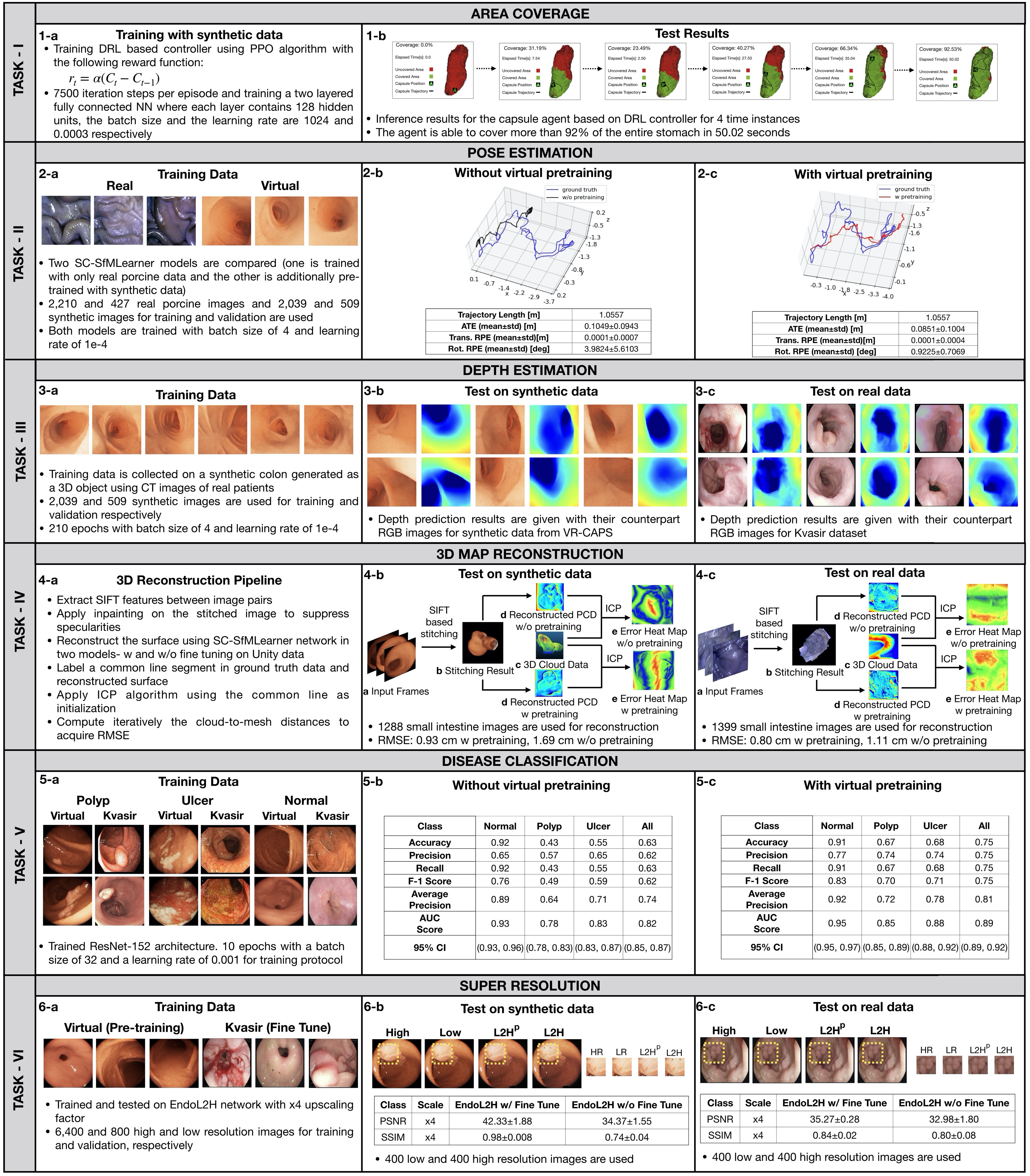}
    \caption{\textbf{Summary of tasks achieved on VR-Caps.} VR-Caps enables to train a Deep Reinforcement Learning (DRL) based controller to autonomously cover the organ. The reward function and training policy for such controller are given in \textbf{1-a} and the resulting covered volumes visually represented with the increasing time steps in \textbf{1-b}. Green and red areas depict the covered and uncovered parts of the organ. For pose estimation, we train two SC-SfMLearner models: one is trained with real porcine data from EndoSLAM dataset and the other one is pre-trained with synthetic data generated on VR-Caps and then fine-tuned with real data. The used data and the information about training policies of pose estimation is given in \textbf{2-a}. In \textbf{2-b} and \textbf{2-c}, we show the comparison results between these two models in terms of predicted trajectory curves with ground truth and qualitative metrics (ATE, Trans. and Rot. RPE). For depth estimation, we train the SC-SfMLearner model with only real data as in \textbf{3-a} and show resulting depth maps for both virtual and real data from Kvasir dataset  in \textbf{3-b} and \textbf{3-c} respectively. For 3D reconstruction, we create point cloud object from synthetic and real data separately as explained in \textbf{4-a}. Then, we compare the reconstructed and ground truth maps in terms of RMSE using ICP algorithm. The pipelines and RMSE results are showed in \textbf{4-b} and \textbf{4-c}, respectively. For disease classification, we created 3 classes: 2 with diseases (Polyps and Ulcer) and 1 without diseases (Normal) and trained two ResNet-152 models (one without pre-training and the other one with virtual pre-training) as represented in \textbf{5-a}. Results in \textbf{5-b} and \textbf{5-c} demonstrates the performance increase in the case of pre-training with VR-Caps data. For super-resolution, as shown in \textbf{6-a} we train EndoL2H network in two scenarios. In first case, we use only real data for training. In second case, we pre-train the network with synthetic data, then fine-tune it with real data. In both cases, we use both synthetic and real data for testing, separately. We show the PSNR and SSIM results of the model with and without pre-training in \textbf{6-b} and \textbf{6-c}}
    \label{fig:results}
\end{figure*}

To demonstrate the effectiveness and usefulness of VR-Caps, we evaluate 6 different tasks in medical applications: area coverage, pose estimation, depth estimation, 3D map reconstruction, disease classification and super resolution.  The following sections describe the evaluation and results for each task in detail, and a brief overview for all these tasks and results is also provided in \Cref{fig:results}.

\begin{figure*}[ht]
\centering
    \includegraphics[width=\textwidth]{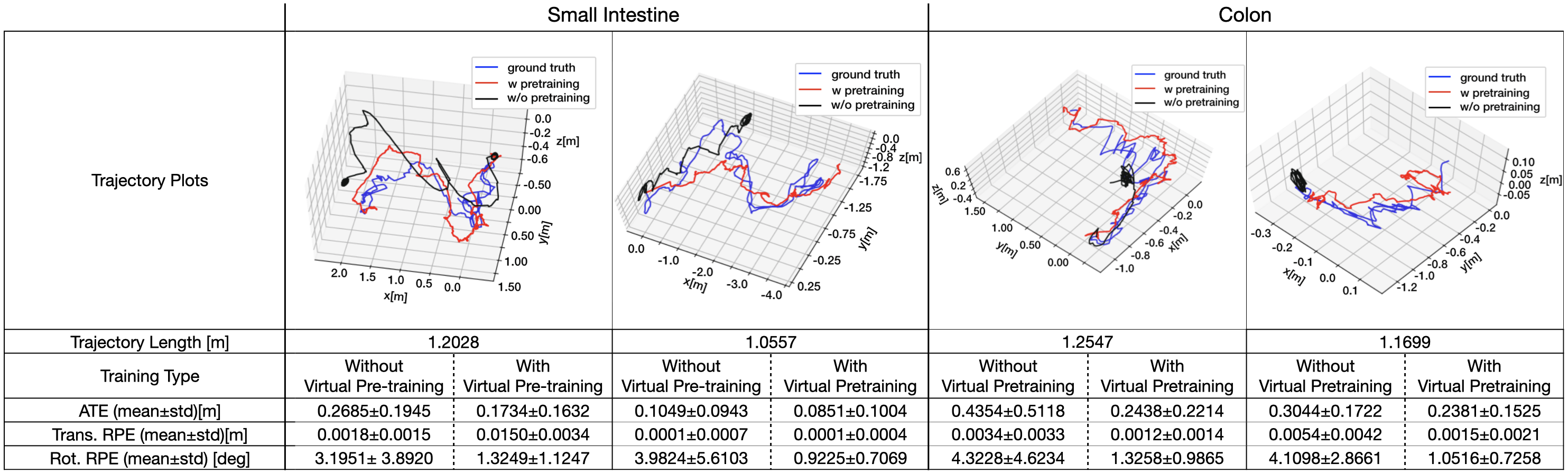}
	\caption{\textbf{Pose estimation results} Unsupervised depth and visual odometry method SC-SfMLearner is used for pose estimation task. We train the network in two different scenarios. In the first case, only with real porcine data from EndoSLAM dataset (without virtual pre-training) is used, whereas in the second scenario, the modeel is pre-trained with synthetic data generated on VR-Caps then fine-tuned with the same set of real porcine data (with virtual pre-training). Both models are tested on the test set of EndoSLAM dataset consisting of 4 different trajectories (2 from small intestine and 2 from colon). Qualitative results are given with trajectory plots of ground truth and aligned predicted trajectories based on the two scenarios. Quantitative results are given with mean and standard deviation ATE (Absolute Trajectory Error), Translation and Rotation RPE (Relative Position Error) scores. Both trajectory plots and numeric results clearly show that in all of the test cases virtual pre-training increases the final pose estimation performance.}
    \label{fig:pose_task}
\end{figure*}

\subsection{Area Coverage}
An estimated 19 million colonoscopies are performed annually in the United States and 6-28\% polyps are missed in routine screenings, indicating a poor sensitivity and accuracy rate for the adenomatous polyps detection \citep{lee2017risk}. Thus, an objective and automated scan of the GI organs is of paramount importance in terms of ensuring a reliable and sensitive endoscopic diagnosis protocol. Such scans are possible with mapping and reconstruction techniques such as SLAM (Simultaneous Localization and Mapping) for solving area coverage \citep{whelan2016elasticfusion, chen2019slam2}. These techniques are widely used in robotics to cope with the challenge of getting a mobile robot to cover the given field in an optimal way. 
The area coverage problem was first addressed by \cite{choset2001coverage} and has become a prominent problem of robotic path planning and exploration in various applications including terrain coverage for autonomous underwater vehicles \citep{hert1996terrain}, robot vacuum cleaner \citep{baek2011integrated} and robotic demining \citep{acar2003path}. To the best of the authors' knowledge, for the first time in literature, we extend the area-coverage problem into endoscopy field. If the explored GI tract area could be systematically and optimally scanned, the disease detection sensitivity of the endoscopist would drastically increase and failed polyp detection rates would drastically decrease. Thus, area coverage has the potential of becoming one of the significant software components of the next generation endoscopic systems. For that purpose, we propose a Deep Reinforcement Learning (DRL) based active control method that has the goal of learning a maximum coverage policy for human organ monitoring within a minimal operation time. We train such control model using the state of the art policy gradient method Proximal Policy Optimization (PPO) \citep{schulman2017proximal} where the model observations are the states of the capsule robot as position and rotation vectors and the output is the discrete set of actions along the $x, y, z$ direction. n this work, we  adopt multi-agent learning paradigm by running three robotic capsule instances inside three different stomach models simultaneously. The reward for the coverage policy is the linear difference coverage difference of all three organs between two consecutive time steps which is stated as:

\begin{equation}
\label{eq:reward}
\color{blue}r_{t} = \alpha \sum_{i=1}^3 (C_{t}^{(i)} - C_{t-1}^{(i)}) 
\end{equation}
where $C_{t}^{(i)}$ represents the percentage of the covered area up until time $t$. It is calculated as the proportion of the vertices that are seen by the capsule camera to the total number of vertices of the $i$th organ instance and $\alpha$ that is a scaling factor for faster convergence and empirically determined as $0.1$. The training is performed with a series of episodes where in each episode the agent performs a set of actions and collects the reward accordingly until the obtained reward converges at a point. In the training session, the length of each episode is set as $7500$ iteration steps and the policy network architecture consists of a three layered fully connected neural network where each layer contains $256$ hidden units, the batch size and the learning rate are $64$ and $3e-4$, respectively. Under these hyper-parameter settings, the agent is trained for $500K$ iteration steps. \Cref{fig:coverage_task} visually demonstrates the covered volume of the test stomach for the inference session for 10 different time instances. Green regions represent covered areas of the inspected organ, whereas red regions show uncovered parts as 3D point clouds projected onto a 2D plane. The arrow indicates the location of the capsule robot and the black line depicts the path that the capsule robot has realized to achieve the coverage. As seen in the figure, the pre-trained agent is able to cover $96\%$ of the inspected stomach instance in $47.62$ seconds. This suggest that a DRL-based autonomous coverage policy method provides a promising and systematic monitoring opportunity for the next-generation capsule endoscopes.

\subsection{Pose and Depth Estimation}
Determining the position of abnormalities is one of the significant challenges in capsule endoscopy, and this requires accurate localization of capsules inside the patient body. Therefore, numerous studies have been conducted in capsule localization such as Radio Frequency localization, Time of Arrival based localization and image-based localization. With the recent developments in image processing techniques, image-based localization methods consisting of pose and depth estimation including Principal Component Analysis (PCA), Vector Quantization (VQ), and machine learning techniques become popular \citep{dey2017wireless}. More recently, numerous deep learning approaches that require large amounts of data have been developed for visual odometry and depth estimation tasks \citep{wang2017deepvo,zhan2018unsupervised,almalioglu2019selfvio}. In this work, the state-of-the-art unsupervised depth and visual odometry deep learning method SC-SfMLearner \citep{bian2019unsupervised} is used to evaluate the effectiveness of synthetic data generated from VR-Caps in pose and depth estimation tasks.
\begin{figure*}[ht]
\centering
    \includegraphics[width=\textwidth]{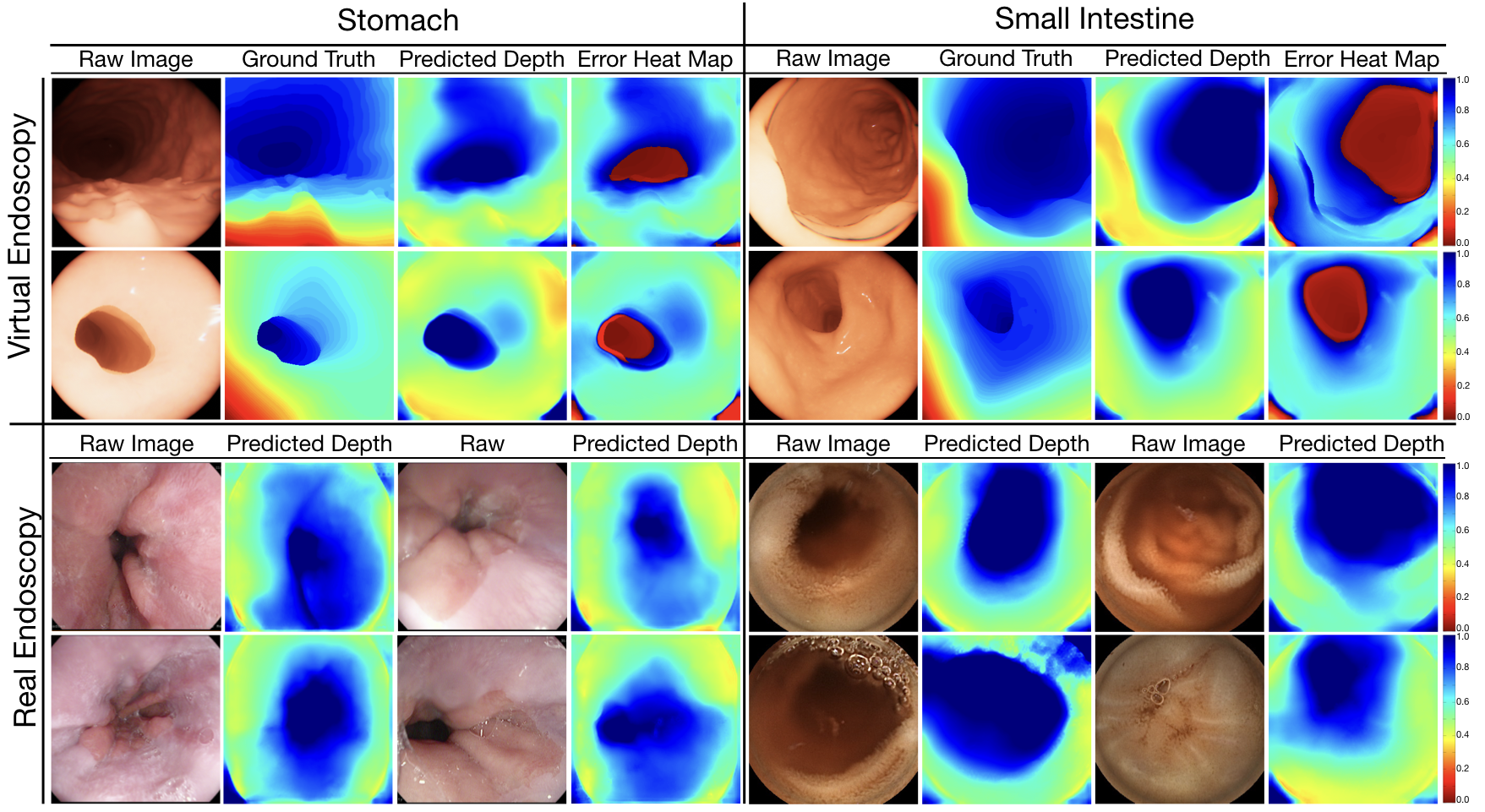}
	\caption{\textbf{Depth estimation results} Depth prediction maps of the unsupervised SC-SfMLearner network trained with only synthetic data generated on VR-Caps. In this figure, we present estimation results for both virtual data and real endoscopy data for stomach and small intestine. As VR-Caps provides ground truth depth for each frame, for virtual test results, ground truth, prediction and error heat maps with their counterpart raw images are provided. For real endoscopy, the trained model is tested on Kvasir dataset for stomach and Redlesion dataset for small intestine. Since there is no ground truth depth information available for these datasets, only depth estimations are provided with their corresponding raw images. It can be seen that unsupervised training with VR-Caps synthetic data qualitatively achieves good performance not only on virtual data but also on real endoscopy data.}
    \label{fig:depth_task}
\end{figure*}
To perform comparative analysis in pose estimation, the SC-SfMLearner network is used in two different training scenarios. In the first scenario, the network is trained using only real endoscopic dataset (EndoSLAM \citep{ozyoruk2020quantitative}) which is acquired from ex-vivo real porcine organ instances and contains endoscopy recordings with 6-DoF ground truth pose. The training is performed with $2,210$ frames, $200$ epochs with a batch size of $4$ and learning rate of $1e-4$. In the second scenario, the same network is first trained with VR-Caps synthetic data, and then fine-tuned on EndoSLAM dataset using the same hyperparameter settings in the first scenario. The synthetic data used for pre-training consists of $2,039$ frames and the network is trained for $210$ epochs with a batch size of $4$ and a learning rate of $1e-4$. Both models are tested on an EndoSLAM test data with four trajectories containing $1135$ and $1525$ images from small intestine and $942$ and $949$ images from colon. Three metrics are used to evaluate the pose estimation performance of these networks: (1) $ATE$ (Absolute Trajectory Error)

\begin{equation}
    \label{eqn:ATE}
    ATE = d(p,q)
\end{equation}
where $d(p,q) = \left((p_1-q_1)^2+(p_2-q_2)^2+(p_3-q_3)^2\right)^{1/2}$ is the Euclidean distance and $p,q \in \mathbb{R}^{3}$ stand for estimated global pose of camera and the ground truth counterparts on the trajectory, respectively. Since both predicted and ground truth trajectories are arbitrarily specified, we use rigid-body transformation in $p$ to a solution that maps the predicted trajectory onto the ground truth trajectory $q$. (2) $RPE_{trans}$ (Translational Relative Pose Error), and (3) $RPE_{rot}$ (Rotational Relative Pose Error) which are stated as:

\begin{equation}
\label{eqn:RPE}
R = (Q_{i}^{-1}Q_{{i+1}})^{-1}\cdot (P_{i}^{-1} P_{{i+1}}) 
\end{equation}
\begin{equation}
\label{eqn:transRPE}
    RPE_{trans} = (R_{0,3}^2 + R_{1,3}^2 + R_{2,3}^2)^{1/2} 
\end{equation}
\begin{equation}
p = \frac{1}{2}(R_{0,0}+R_{1,1}+R_{2,2}-1) 
\end{equation}
\begin{equation}
\label{eqn:rotRPE}
RPE_{rot} = \arccos(\max(\min(p,1),-1))
\end{equation}
where $R \in {\rm I\!R} ^{4x4}$ represents relative pose error. $P$ and $Q$ stand for predicted and ground truth pose, respectively. In \Cref{fig:pose_task}, the results of two training scenarios on the EndoSLAM test set are represented qualitatively with aligned trajectory plots, and quantitatively with the aforementioned metrics. Both qualitative and quantitative results clearly show that in all cases pre-training with synthetic data improves the final pose estimation performance.

\begin{figure*}
    \includegraphics[width=\textwidth]{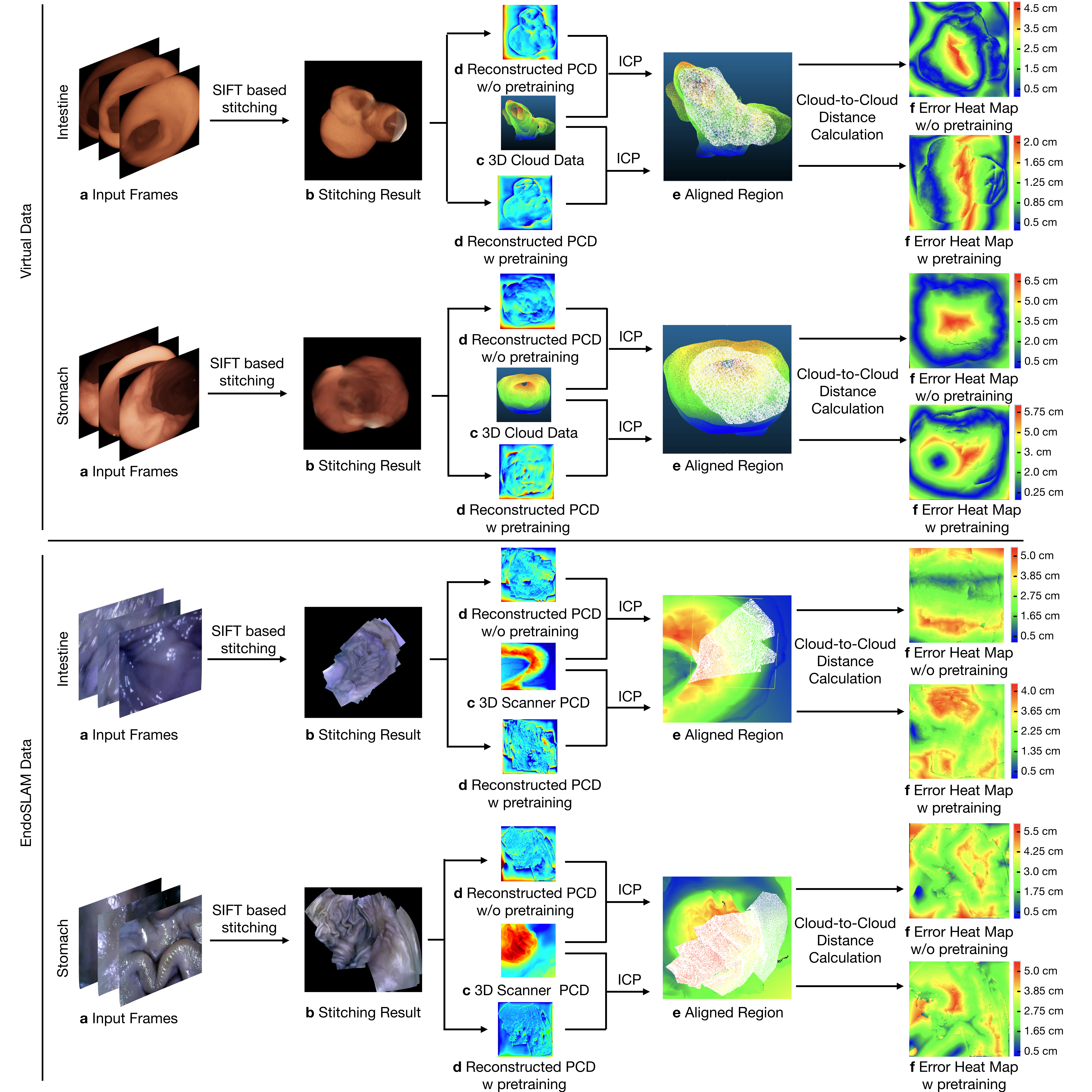}
	\caption{\textbf{3D Reconstruction results}. 
	This figure illustrates the 3D reconstruction results for using both VR-Caps and EndoSLAM data. To show the impact of using synthetic data generated by VR-Caps, results are tested on two cases where the virtual data is not used (w/o pre-training) and the virtual is used for pre-training (w pre-training).  \textbf{a)} The unordered input frames of small intestine and stomach from VR-Caps and EndoSLAM are given as input to Scale Invariant Feature Transform.  \textbf{b)} The final stitched image is created by aligning and multi-blending all input images. Specularities on stitched image are inpainted. \textbf{c)} The point cloud data in .ply format for synthetic data is provided by VR-Caps while the point cloud is obtained using 3D scanner in EndoSLAM dataset. \textbf{d)} The depth map of inpainted image is predicted based on two models of SC-SfMLearner network (w and w/o virtual pre-training). \textbf{e)} The matched area between ground truth data and reconstructed surface is formed by using ICP (Iterative Closest Point) algorithm. \textbf{f)} The cloud-to-mesh/cloud-to-cloud distances are displayed in the form of heatmap. The bar shows the root mean square error in unit of [cm]. In the case of VR-Caps data is not used for training, the RMSEs for small intestine and stomach are 1.69 and 2.45 [cm] and 1.11 and 1.30 [cm] for synthetic data and real porcine data respectively. Whereas, for the model pre-trained with virtual data, the RMSEs for small intestine and stomach are 0.93 and 2.26 [cm] and 0.80 and 0.92 [cm] for synthetic and real data respectively. The difference reveals the effectiveness of using VR-Caps in 3D reconstruction by showing better alignment with the ground truth data.}
    \label{fig:3dReconstructionFigure}
\end{figure*} 

To evaluate the depth estimation performance, we test the model trained with only synthetic data and test on both synthetic (virtual stomach and small intestine from VR-Caps) and real endoscopy images (Kvasir dataset for stomach and Redlesion \cite{coelho2018deep} for small intestine). In \Cref{fig:depth_task}, we show sample depth maps for these test cases. Since VR-Caps is able to provide ground truth depth for every frame generated, error heat maps are given to depict the estimation performance of the model. As there is no ground truth depth available for real endoscopy datasets, we present only the qualitative results. The results suggest that in a cross-dataset setting, deep learning models trained on virtual images have sufficient generalization power on depth estimation task.

\subsection{3D Reconstruction}
\label{ssec:3drec}
In endoscopy and colonoscopy, doctors are still challenged by some limitations such as the constrained field-of-view and the uncertainty of endoscope location inside the organ under inspection. Determining the location of lesions with respect to complete view of inspected organs is crucial for surgeons to plan treatment procedure in early stages of the malignancy and this can be possible by reconstructing the 3D shape of the whole organ based on visual feedback from endoscope camera \citep{widya20193d}. Existing works have proposed solutions to reconstruct the 3D shape of organs (e.g., colon, liver, and larynx) based on camera pose acquired by endoscope videos. These solutions span shape-from-shading, visual SLAM, and structure-from motion. Some reconstruction artifacts due to repeating patterns may be removed using periodic plus smooth decomposition \citep{mahmood20152d}. 

\begin{algorithm}[h]
\caption{3D Reconstruction and Evaluation Pipeline}
\label{algo:3D}
\begin{algorithmic}[1]
    \State Extract SIFT features between image pairs
    \State Find $k$-nearest neighbours for each feature using a  $k$-d tree
    \For{each image}
        \State (i) Select $m$ candidate matching images that have the  most number of corresponding feature points
        \State (ii) Find geometrically consistent feature matches using RANSAC to solve 
        for the homography between pairs of images. 
    \EndFor 
    \State Find connected components of image matches 
    \For{each connected component}
        \State (i) Perform bundle adjustment for connected components in image matches
        \State (ii) Render final stitched image using multi-band blending
    \EndFor 
\State Apply inpainting on the stitched image to suppress specularities
\State Reconstruct the depth surface using SC-SfMLearner network
\State Label a common polygon segment in ground truth data and reconstructed surface
\State Apply ICP algorithm using the common polygon as initialization
\State Compute iteratively the cloud-to-cloud distances to meet RMSE termination criteria
\end{algorithmic}
\end{algorithm}
In this work, we use a hybrid 3D reconstruction technique described in \Cref{algo:3D}. The fundamental steps are Otsu threshold based reflection detection, inpainting-based reflection suppression, feature matching, tracking based image stitching and non-Lambertian surface reconstruction. Feature point correspondences between frames are created using SIFT feature matching and RANSAC based pair elimination \citep{autostitch2,autostitch}. Then the surface reconstruction is performed using the depth images obtained from SC-SfMLearner network. To align the ground truth data and reconstructed surface, the ICP approach is employed after manually labelling a common polygon segment. To demonstrate the impact of synthetic data generated on VR-Caps, we evaluate the reconstructed surfaces using depth maps obtained from two different networks (with and without pre-training) as seen in \Cref{fig:3dReconstructionFigure}. We compare the predicted reconstructions for both synthetic and real data containing 3D ground-truth depth maps. In terms of RMSE values, reconstructed surfaces of observed organ instances obtained from pre-trained model demonstrates better alignment with corresponding ground truth data. Such that for the pre-trained model, the RMSE scores for small intestine and stomach are 0.93 and 2.26 [cm] and 0.80 and 0.92 [cm] in the case of synthetic data and real data respectively. Whereas, for the model where VR-Caps data is not used, the RMSE scores for small intestine and stomach are 1.69 and 2.45 [cm] and 1.11 and 1.30 [cm] for synthetic data and real porcine data respectively. Cloud-to-cloud distances for reconstruction processes are provided in \Cref{fig:3dReconstructionFigure} as heatmap representation.
\begin{figure}[!h]
    \includegraphics[width=\columnwidth]{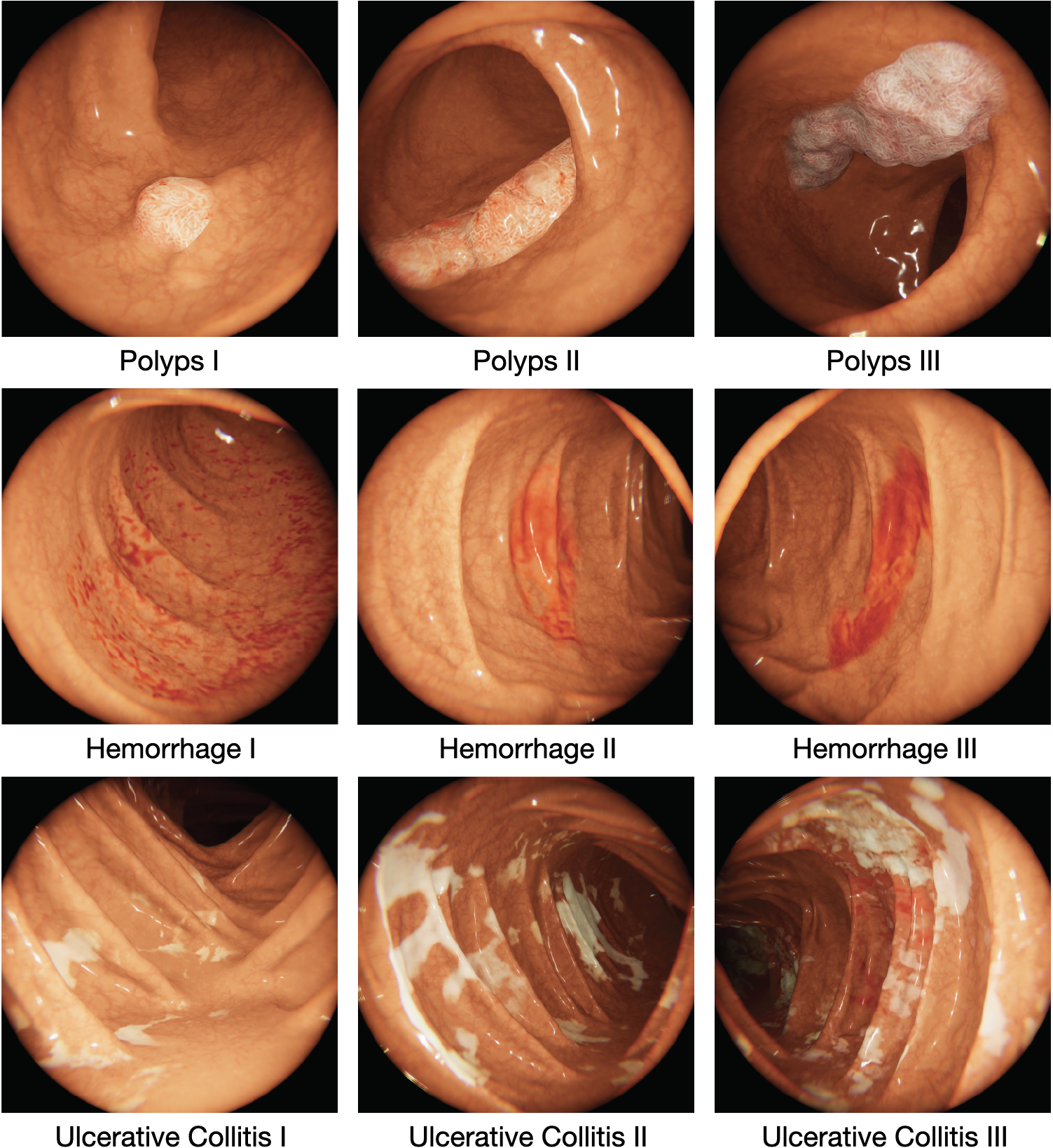}
	\caption{\textbf{Disease classes} 3 diseases are generated in VR-Caps and instances with various shapes, sizes and severity levels are shown. Hemorrhage and Ulcerative Collitis are created based on the real endoscopy images mimicking the abnormal mucosa texture. Polyps are reconstructed based on CT scans as they are distinctive in topology.}
    \label{fig:diseaseFigure}
\end{figure}
\begin{figure*}[!ht]
    \includegraphics[width=\textwidth]{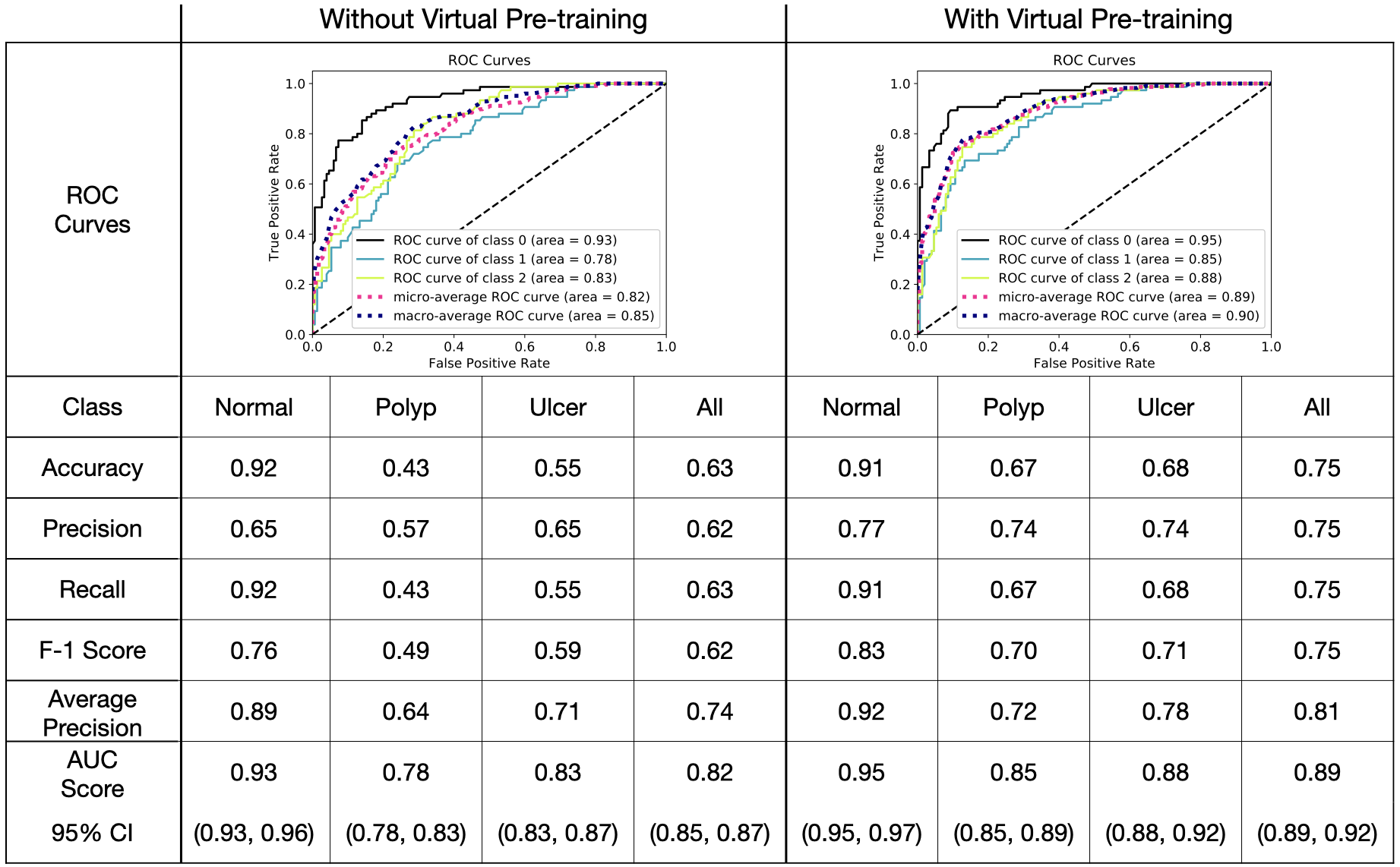}
	\caption{\textbf{Disease classification results} This figure demonstrates the comparative results for two scenarios in disease classification task where the classes are Normal, Polyp and Ulcer. In the first scenario, ResNet-152 CNN architecture is trained on real endoscopy data (Kvasir) which consists of 425 images per class (with virtual pre-training). Whereas, in the other scenario, the same network is initially pre-trained with synthetic data generated using VR-Caps which contains 2940, 3400, 2374 images for Normal, Polyps and Ulcer, respectively. The pre-trained model is fine-tuned with the same training data used in the first scenario. Both models are tested on Kvasir test dataset containing 225 real endoscopy images and numeric results as well as ROC curves for each case are presented. The model with virtual pre-training significantly improves the classification performance compared to the model without pre-training in all metrics except a minor decrease in accuracy and recall for the Normal class.}
    \label{fig:classification_task}
\end{figure*}
\subsection{Disease Classification}
\label{sec:disease-classification}
In order to support diagnostics procedure and help physicians, the newer image recognition methodologies (i.e., computer aided systems) have been applied to many aspects in the medical field. Among these methods, CNNs have been widely used particularly for automated lesion segmentation in endoscopy images \citep{wang2020multi} and detecting and diagnosing colorectal diseases from capsule endoscopy videos \citep{takiyama2018automatic,mahmood2019polypseg}. To provide synthetic data that can be used in training of these CNN models, in VR-Caps environment, we generate instances of 3 different disease classes with varying severity grades. Diseases that we introduce in the environment are polyps, ulcerative colitis, and hemorrhoids with various shape, size and grades. CT-scans from TCIA dataset contains several polyps of different shapes and variations that allows for reconstructing these instances inside the virtual 3D organ models. On top of the volumetric reconstruction, we assign texture on these virtual polyp instances using extracted textures from Kvasir dataset. In total, 4 polyp instances are generated with several variations in shapes. By assigning texture frames obtained from 5 different polyp instances with different severity levels in the Kvasir dataset, 20 polyp instances with varying shape and textures are created. As a second disease class, instances from 3 different grades of ulcerative colitis are generated. Then, by applying these textures with 20 different shapes and size of areas to the various segments that are created during UV mapping, we create 60 instances for ulcerative colitis disease class, in total. Hemorrhoid bleeding disease class, is created using 4 different grade levels taken from the Kvasir dataset and by applying them in 10 different shape and size of areas to the segments, we get 40 instances with varying amount of bleeding. Examples of these diseases with varying grades can be seen in \Cref{fig:diseaseFigure}.
\begin{figure*}[h]
    \includegraphics[width=\textwidth]{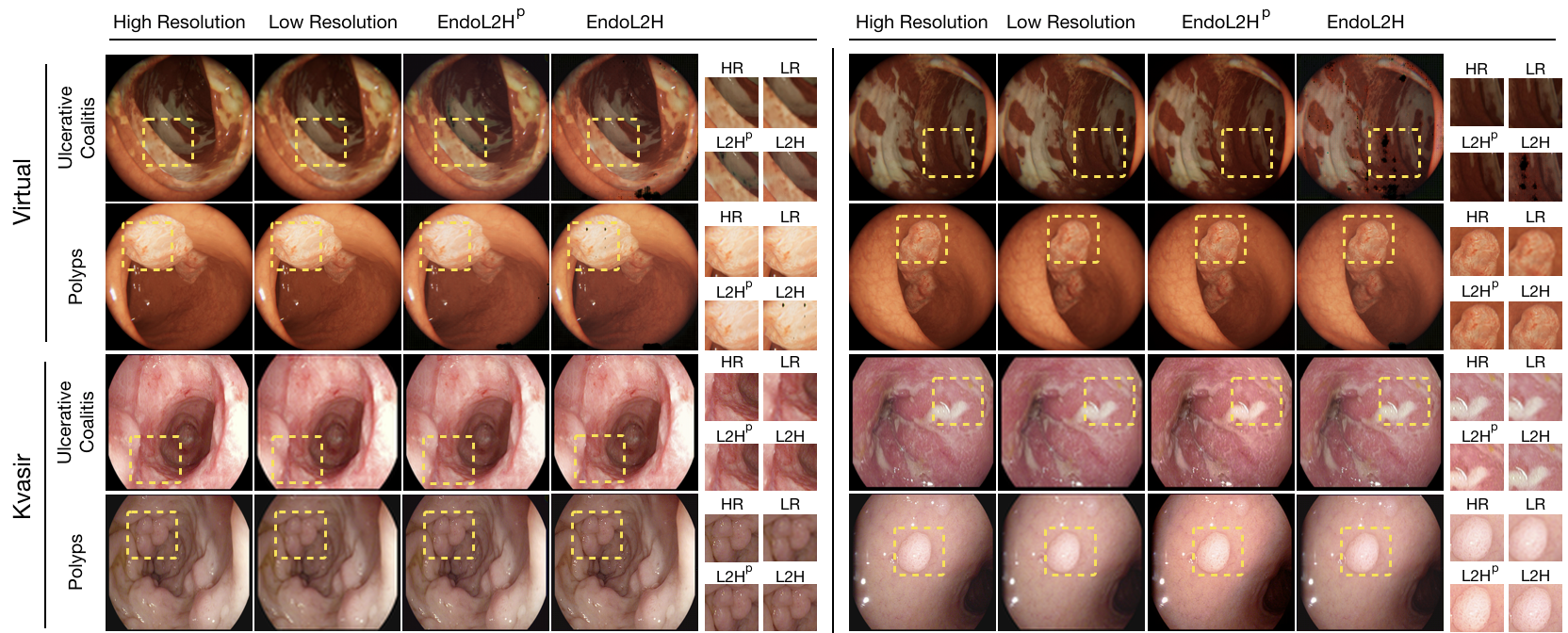}
	\caption{\textbf{Results on superresolved diseased regions}. Super Resolution results of ulcerative coalitis and polyps disease classes for both VR-Caps and Kvasir datasets with EndoL2H algorithm on 4x scale factor. The diseased regions are zoomed and cropped inside the yellow squares. $EndoL2H$ indicates that the model is trained with real Kvasir dataset whereas $EndoL2H^p$ is the pre-trained version of $EndoL2H$ with virtual data.  
	PSNR and SSIM metrics are used to evaluate the performances of these two protocols.}
    \label{fig:SuperResolutionFigure}
\end{figure*}
To exemplify the use case of VR-Caps in disease classification, we train a ResNet-152 model for solving image classification with three classes: Polyp, Ulcer, and Normal. The hemorrhoid class is excluded since it is not present in the benchmark dataset (Kvasir). 
The VR-Caps dataset for disease classification task consists of 8714 image frames in total where each class Normal, Polyp, and Ulcer contains 2940, 3400, 2374 frames respectively. A comparative analysis based on two scenarios proves that VR-Caps provides potentially useful synthetic data for classification tasks. For the first scenario (without virtual pre-training), the ResNet-152 CNN architecture is trained on real endoscopy data (Kvasir) which consists of 1275 images where each of the three classes have equal number of frames which is 425. In the second scenario (with virtual pre-training), the same network is first pre-trained on virtual data generated by VR-Caps, then fine-tuned with the same training data used in the first scenario. The same hyperparameter settings are used for all trainings in both scenarios where the number of epoch, batch size, and the learning rate are $10$, $32$, $0.001$, respectively. Both scenarios are tested on a common Kvasir test set which contains 225 frames for each class. The classification results for both cases are presented in \Cref{fig:classification_task}. The model with virtual pre-training significantly outperforms the model without pre-training in all metrics including accuracy, precision, recall, average precision and AUC (Area Under a Curve) score with 95\% CI (Confidence Interval). For instance, virtual pre-training with VR-Caps dataset increases the percentage of the f-1 scores of Polyp and Ulcer classes by 21\% and 11\%, respectively. Only a minor decrease is observed in accuracy and recall metrics for the Normal class.

\subsection{Super Resolution}
Current practices of capsule endoscopy suffer from low resolution images  due to low lighting and limited power. The poor image quality, therefore, brings difficulty in accurate diagnoses. Due to size and cost limitations in imaging sensors of the capsule cameras, methods for super-resolution (i.e., generating high resolution images from low resolution frames) have become an important clinical problem for building more hardware-efficient imaging devices. \citep{park2019recent}. 

To investigate how simulated data can be used to solve this problem, we benchmark the VR-Caps using the Deep Super-Resolution for Capsule Endoscopy (EndoL2H) network  \citep{almalioglu2020endol2h}. The defined task is to superresolve the input frames with a resolution of $256\times256$ pixels to a resolution of $1024\times1024$, making an upscaling of 4x in total. As validation, we devise two ablation experiments. In our first experiment, we train and validate EndoL2H network using real frames from Kvasir dataset for $100$ epochs and test on both synthetic and real dataset, with $6400$ real frames for training, $800$ real frames for validation and $800$ frames from VR-Caps and Kvasir dataset for testing are used. In the second experiment, we train and validate the network with synthetic frames for $50$ epochs, then fine-tune using Kvasir dataset for another $50$ epochs. From each of VR-Caps and Kvasir dataset, $6400$ frames  for training, $800$ frames for validation and $800$ frames for testing are used, respectively. To evaluate the super resolution performance, Peak Signal-to-Noise Ratio (PSNR) and Structural Similarity Index (SSIM) metrics are used. 
By pre-training the model on VR-Caps data, better PSNR and SSIM values are obtained with an increase in the scores from 34.37 \& 0.74 to 42.33 \& 0.98 for virtual and from 32.98 \& 0.80 to 35.27 \& 0.84 for real test data, respectively. 
In addition, super resolution effect on diseased regions of organ instances for both synthetic and real data is qualitatively represented in \Cref{fig:SuperResolutionFigure}.

\section{Conclusions and Future Work}
\label{sec:discussionandconclusion}
In this work, we introduced a virtual endoscopy environment, VR-Caps, that is capable of generating fully-labeled and realistic synthetic data that is consistent with the topology and texture of real organs for control, navigation, and machine vision related tasks, which may contribute to significant improvements in several medical imaging and device control applications. We have illustrated the effectiveness of the synthetic data generated by our environment on state-of-the-art methods applied to 6 different tasks. These tasks showed that data provided by VR-Caps increases the performance of deep learning models that can enhance the effectiveness of endoscopy by providing solutions to common technical problems faced during surgical or diagnostic procedures. For endoscopic control and vision related tasks, we believe that we reduced the gap between simulation and real-world domains which is essential in transferring learned skills to real-world robotic scenarios. 
Besides various sim-2-real scenarios, we additionally addressed the autonomous area coverage problem and showed a promising coverage learning strategy for endoscopic systems. As future work, we will continue working on the coverage problem and evaluate the validity of our approach in real-world scenarios. We also plan to extend the environment to simulate interaction with flexible or tethered double balloon endoscopes and to include additional disease classes and modalities for control and vision tasks. We also plan to leverage the environment to study the effect of capsule design choices on the performance of the various algorithms (e.g., conducting depth estimation test with a stereo camera or coverage task with a 360\degree camera design). 

\section*{Declaration of Competing Interest}
The authors declare that they have no known competing financial interests or personal relationships that could have appeared to influence the work reported in this paper.

\section*{Acknowledgements}
Mehmet Turan, Kagan  Incetan, Ibrahim O. Celik, Abdulhamid Obeid, Guliz I. Gokceler, Kutsev  B.  Ozyoruk  are especially grateful to Technological Research Council of Turkey (TUBITAK) for International Fellowship for Outstanding Researchers grant.

%\clearpage
\bibliography{references}
%\clearpage
\pagenumbering{gobble}

\end{document}